\pdfoutput=1
\documentclass[11pt]{article}

\usepackage[]{acl}

\usepackage{times}
\usepackage{latexsym}

\usepackage[T1]{fontenc}
\usepackage[utf8]{inputenc}

\usepackage{microtype}

\usepackage[inline]{enumitem}

\usepackage{amsfonts}

\usepackage{graphicx}

\usepackage{multirow}

\usepackage{booktabs}

\usepackage{tabularx}
\usepackage{makecell}

\usepackage{caption}
\usepackage{subcaption}
\usepackage{amsmath,amssymb}
\usepackage{mathtools}
\usepackage[thinc]{esdiff}

\newcommand{\pfone}{AMUE }
\newtheorem{define}{Definition}
\newcommand{\appropto}{\mathrel{\vcenter{
  \offinterlineskip\halign{\hfil$##$\cr
    \propto\cr\noalign{\kern2pt}\sim\cr\noalign{\kern-2pt}}}}}

\title{On the Economics of Multilingual Few-shot Learning: Modeling the Cost-Performance Trade-offs of Machine Translated and Manual Data}

\author{Kabir Ahuja\textsuperscript{1} \quad Monojit Choudhury\textsuperscript{2} \quad Sandipan Dandapat\textsuperscript{2}\\
\textsuperscript{1} Microsoft Research, India \\
\textsuperscript{2} Microsoft IDC \\
{\tt \small \{t-kabirahuja,monojitc,sadandap\}@microsoft.com}
}

\begin{document}
\maketitle
\begin{abstract}
%Massively Multilingual Language Models have recently been shown to be surprisingly effective in zero-shot transfer. Naturally, the presence of the task specific data in the target language improves over the zero-shot performance, often substantially. Two common ways of obtaining such data in the target language that are explored in literature are i) Translating the labelled data present in the Pivot Language, ii) Obtaining few-shot examples for the task in target language, and have been shown to be quite effective towards improving the zero-shot performance. However, the study on the comparison of the performance trade-offs between these two strategies has been limited. Adding to that  there is also a cost trade-off attached to these i.e. \textit{translation costs} and \textit{annotation costs}. 
Borrowing ideas from {\em Production functions} in micro-economics, in this paper we introduce a framework 
to systematically evaluate the performance and cost trade-offs between machine-translated and manually-created labelled data for task-specific fine-tuning of massively multilingual language models. We illustrate the effectiveness of our framework through a case-study on the TyDIQA-GoldP dataset. One of the interesting conclusions of the study is that if the cost of machine translation is greater than zero, the optimal performance at least cost is always achieved with at least some or only manually-created data. To our knowledge, this is the first attempt towards extending the concept of production functions to study data collection strategies for training multilingual models, and can serve as a valuable tool for other similar cost vs data trade-offs in NLP.
\end{abstract}

%Introduction Section
% 1. Intro (1 Page)
%     - SCI: Zero-Shot is reasonable but few-shot and translation gives better performance. No study on tradeoff especially when we are mixing the two. People might just use one. Related work should come here.
%     - Problem Statement: Given a budget  a pretrained language model and corpus of data for the task in 1 or few pivot languages, what is the maximum acheivable accuracy in a target language through fine-tuning on pivot language data + translation data + fresh labelled data in target language, where the latter two will have some costs attached to them.
%     - Solving the above problem or analyze the issue one has to understand the exact nature of tradeoffs between these 4 factors.(There are also other factors like choice of pivot languages, model types, keeping the constant).
%     - Microeconomics has the concepts (pfs, isoqs etc) which have one to one mapping with what we are trying to do here. We introduce the framework and present a case study on TyDIQA.
%     - Contribution: 1. Framework, first use of this and can be applied to a multitude of NLP problems where such tradeoffs are commonly seen. 2. Fitting these performance functions 3. Learnings from Case study. As far as we know these two tradeoffs are studied and microeconmics concepts are concepts are applied.

\section{Introduction}
Transformer based Massively Multilingual Language Models (MMLMs) such as mBERT \cite{devlin-etal-2019-bert} , XLM-RoBERTa \cite{conneau-etal-2020-unsupervised} and mT5 \cite{xue-etal-2021-mt5} are surprisingly effective at zero-shot cross-lingual transfer ~\cite{pires-etal-2019-multilingual, wu-dredze-2019-beto}. However, while zero-shot transfer is effective, often the performances across different languages is not consistent. Low-resource languages \cite{wu-dredze-2020-languages} and the languages that are typologically distant from the pivot language \cite{lauscher-etal-2020-zero} are known to benefit the least from zero-shot transfer, which can often be mitigated by using target-language specific labelled data for the task during fine-tuning. 

One common approach for collecting such data in the target language is to translate the training data available for the pivot-language to the target by using an off-the-shelf Machine Translation (MT) system. This is usually referred to as the \textit{translate-train} setup ~\cite{huetalXTREME, turcetalRevisiting}. Few-shot transfer is another alternative; as shown by \citet{lauscher-etal-2020-zero}, a few labelled examples in the target language, that can be obtained cheaply, can lead to substantial improvements over the zero-shot performance. 

However, there has not been much work on comparing the performance across these two strategies. In one such study, \citet{huetalXTREME} compare the performance of \textit{translate-train} with \textit{few-shot} transfer on TyDIQA-GoldP~\cite{clark-etal-2020-tydi} dataset, but they only evaluate the few-shot case with 1000 examples, which does not provide any insight into how the performance varies with increasing dataset sizes for these two approaches. Additionally, there are trade-offs related to the data acquisition costs as well. The cost per training instance is expected to be much smaller for an MT-based approach than manual translation or labeling of examples. However, depending on the nature of task, language, and quality of the MT output, the amount of data required to achieve the same performance through these two approaches can be drastically different. More importantly, fine-tuning the MMLMs with a combination of the data from the two strategies could be the cheapest alternative for achieving a target accuracy, which, to the best of our knowledge, has not been explored yet.

Inspired by the above observations and gaps, in this paper, we ask the following question: Given a pre-determined budget to fine-tune a multilingual model on a task for which some data is available in a pivot language, what is the best achievable accuracy on a target language by (a) training the model on the pivot-language data, (b) different amounts of machine-translated and (c) manually-collected data in the target language. Solving this requires an understanding of the exact nature of the performance and cost trade-offs between the two kinds of target language datasets and their relative costs of acquisition, apart from factors such as the amount of pivot language data, the task, the MMLM, and the languages concerned.

This problem of modeling and measuring the trade-offs between different input factors and their costs is well-studied in the field of micro-economics. A sophisticated machinery has been developed in the form of {\em Production Functions} and allied analytical methods~\cite{miller2009input,cobb1928theory}, in order to solve the following generic problem: with the best available technology, how are the inputs to a production process (eg.  \textit{Labor} and \textit{Capital}) related to its output, that is the quantity of goods produced. In this paper, we adapt this framework to address the aforementioned question of MMLM fine-tuning trade-offs.  

The key contributions of our work are \textit{threefold}. \begin{enumerate*}
    \item[1.] We extend the idea of production functions to {\em performance functions} that model the relationship between input data sizes and performance of a system; we propose a possible analytical form for this function and derive the performance trends and optimal data collection strategies under fixed costs. 
    \item[2.] We illustrate the usefulness of this framework through a case study on a Q\&A task -- TyDIQA-GoldP \cite{clark-etal-2020-tydi} and systematically study the various trade-offs for 8 languages. 
    \item[3.] Our study provides several important insights such as (a) sample efficiency of manual data is at-least an order of magnitude higher when compared to translated data,  (b) if the cost of MT data creation is non-zero, then the optimal performance under a fixed budget is always achieved with either only manually-created data or a combination of the two; (c) the ratio of the two datasets for the least cost combination usually remains constant at different levels of performance. 
\end{enumerate*}

To the best of our knowledge, this is the first work that applies the idea of production functions to analyze the cost-performance trade-offs of MMLM fine-tuning. The proposed framework can be extended to a multitude of NLP problems where the trade-offs similar to the ones discussed above, are common (e.g., pre-training vs. fine-tuning data). 
To encourage reproducible research, we have made our code, the performance data, and a detailed list of the results publicly available \footnote{\href{https://github.com/kabirahuja2431/PerformanceFunctionAnalysis}{github.com/kabirahuja2431/PerformanceFunctionAnalysis}}.

\section{Theoretical Foundations}

One of the foundational pillars of neoclassical economics is the idea of {\em Production Functions}. Simply put, a production function is a mathematical formalization of the relationship between the output of a firm (industry, economy) and the inputs that have been used in obtaining it~\cite{khatskevich2018production,miller2009input}. A multi-factor production function is defined as a map 
  \begin{equation}
    Q: \mathbf{x} \rightarrow f(\mathbf{x}), \quad \forall \mathbf{x} \in \mathbb{R}^{+n}
   \end{equation}
  where $Q \in \mathbb{R}^+$ is the quantity of output, $n$ is the number of the inputs, the non-negative function $f$ is continuously differentiable for all $\mathbf{x} = (x_1, \dots, x_n)$ when $x_i \in \mathbb{R}^+$. A sophisticated and extensive set of analytical machinery has been developed over the years in microeconomics theory that allows one to closely model and analyze not only the relationship between the inputs and outputs\footnote{Production functions can also be defined when there are $m$ outputs i.e. $Q \in \mathbb{R}^m$ and $f : \mathbb{R}^n \rightarrow \mathbb{R}^m$.}
  of a firm, but also the interdependence between the inputs (i.e., $x_i$s). Thus, one can efficiently compute and clearly visualize the various trade-offs and optimal configurations of the production system.
  
  %This can be best illustrated by taking the example of $L$ units of labor and $K$ units of capital as the input factors and $Q = f(L,K)$, as the quantity of a product produced by a firm. $L$ and $K$, can be \textit{perfect substitutes}, i.e., the same level of output $Q$ can be produced by substituting the units of capital and labor at a constant rate. The production function in such a case takes a linear form, i.e.,
 % \begin{equation}
 %     Q = \frac{L}{a_l} + \frac{K}{a_k}
  %\end{equation}
  %meaning a single unit of output can either be produced by $a_l$ units of labour {\em or} $a_k$ units of capital. On the other end of the extreme are the \textit{perfect complements} which must be used in fixed proportions to produce a given quantity of product. The production function then is of the form 
 % \begin{equation}
 %     Q = \min(\frac{L}{a_l}, \frac{K}{a_k})
 % \end{equation}
 %That is to say, in order to produce one unit of product $a_l$ units of labour {\em and} $a_k$ units of capital are required. This is also known as the {\em Leontief} production function~\cite{miller2009input}. Most practical scenarios fall in between these two extremes and are called \textit{imperfect substitutes}.
  
  Production functions have been extensively used to model and study systems as diverse as education~\cite{bettinger2020does,bowles1970towards}, environment~\cite{lu2019impacts,halicioglu2018output}, sustainability~\cite{yankovyi2021modeling}, cognition~\cite{todd2003specification} and of course, different types of industries~\cite{husain2016test,batiese1992frontier}. Along similar lines, in this work we develop the concept of \textit{Performance Function} that models the performance of an MMLM given the amount of translated and manually labeled data. In this section, we begin by formalizing the notations and defining some key concepts from micro-economics, appropriately adapted to our context. Then we present the functional form of the performance function, and discuss certain practical constraints and assumptions that we will make in our formulation.

\subsection{Notation and Definitions}
Consider a multilingual model $\mathcal{M}$ pre-trained on a set of languages $\mathcal{L}$, which is to be fine-tuned for a task $\mathfrak{T}$, for which $P$ labelled examples are available in a pivot language $p \in \mathcal{L}$. Some or all of the $P$ pivot language examples can be automatically translated to a target language $l \in \mathcal{L}$ through an MT system to obtain $T (\le P)$ examples. Further, let $M$ be the amount of  examples for $l$ that have been labelled or translated manually.  

\begin{define}
{\bf Performance Function}, $\Pi = \pi(T, M | l, p, P, \mathcal{M}, \mathfrak{T})$, denotes the {\em best} possible performance (as per the current state-of-the-art) of a system in language $l$ for a task $\mathfrak{T}$, that has been built on top of a pre-trained MMLM $\mathcal{M}$, $P$ labelled examples in language $p \ne l$, $T$ translated examples by an MT system, and $M$ manually created examples.
\end{define} 

Here, $\Pi \in [0,1]$ is any appropriate and accepted measure of  performance, such as accuracy or F1-score. To simplify the notation we will often drop the given conditions from the equation and denote $\Pi = \pi(T, M)$. The conditional factors, whenever not obvious from the context, will be explicitly stated. Note that $T$ and $M$ are respective equivalents of $K$ and $L$ of the neoclassical Capital-Labor production functions. Capital investment in technology or mechanization is similar to {\em machine}-translated data, whereas manual dataset creation would require investment on labor. 

\begin{define}
{\bf Sample Efficiency} (or 
Marginal Product) of a given source of data $X$ ($T$ or $M$),  $\Psi = \psi(X)$, is the partial derivative of the performance function $\Pi$ w.r.t. the input factors i.e. the change in performance value for a unit change in manual or translated data.
\end{define}

\begin{align*}
    \psi(T) &= \frac{\partial \pi(T,M)}{\partial T}\\
    \psi(M) &= \frac{\partial \pi(T,M)}{\partial M}
\end{align*}

Here, we denote $\psi(T)$  and $\psi(M)$ as the sample efficiencies of translated and manual data respectively. 

\begin{define} 
{\bf Total cost of operation} (or simply the cost),
$\kappa(T,M) = \kappa_t(T) + \kappa_m(M)$, is the total cost of procuring translated and manually created datasets for $l$ for the task $\mathfrak{T}$.  
\end{define}
We further assume that the translation and manual collection costs are scalar multiples of the unit costs, i.e. $\kappa_t(T) = c_tT$ and $\kappa_m(M) = c_mM$, where $c_t > 0$ is the cost of translating a single example from $P$ into language $l$ automatically, while $c_m > 0$ is the cost of collecting one training example in $l$ manually. Therefore,
\begin{equation}
    C = \kappa(T,M) = c_tT + c_mM 
\end{equation}

Usually, $c_m > c_t$. Also, note that we are ignoring the costs of pivot data collection and computational costs of pre-training and fine-tuning, partly because we are interested in studying the trade-off between $T$ and $M$. Also, $P$ is useful for any target language, and therefore, the amortized cost of creating $P$ tends to zero as the number of target languages increases. Similarly the amortized cost of pre-training tends to zero as the number of tasks grows. The task-specific training cost is proportional to training data-size, $P+T+M$, and therefore, can be partially consumed in $c_t$ and $c_m$. 

%data collection cost, pivot data collection cost and training costs, i.e., $C = \texttt{COST-TRANS}(T) + \texttt{COST-MAN}(M) + \texttt{COST-PIV}(P) + \texttt{COST-TRAIN}(\mathcal{M}, P, T, M)$. For our analysis, we assume To simplify the analysis and put main focus on the trade-offs between Translated and Manual data, we assume the pivot data cost and training costs to be 0, giving 
\begin{figure}
    \includegraphics[width=0.48\textwidth]{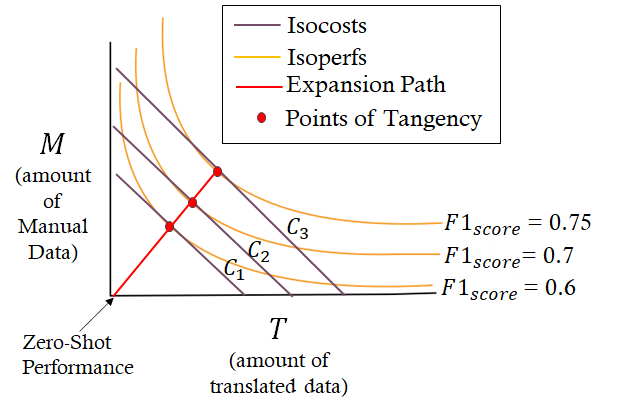}
    \caption{Hypothetical T-M diagram illustrating \textit{isoperfs}, \textit{isocosts}, \textit{points of tangency} and \textit{expansion path}.}
    \label{fig:exp_path}
\end{figure}

\begin{define}
{\bf Isoperf} curves are the contours of the performance function that represent the relationship between $T$ and $M$ for a fixed performance value $\Pi_c$.
\\
\end{define}

\begin{define}
{\bf Isocost} curves are the contours of the cost function that represent the different possible combinations of $T$ and $M$ that result in equal overall costs.
\end{define}

Both isoperfs and isocosts are drawn on a T-M diagram (K-L diagram in micro-economics), which is illustrated in Fig.~\ref{fig:exp_path}. The x and y axis represent the input factors T and M, respectively. The orange curves are the hypothetical isoperfs, known as {\em isoquants} in economics. As the name suggests, each point on these curves represents T-M combinations that result in the same ({\em iso}) performance ({\em perf}), denoted in the diagram by $\Pi_1$, $\Pi_2$, etc. Intuitively, it can be seen that two isoperfs never intersect; as we move towards right and up, $\Pi$ increases because either $T$ or $M$ or both increase. Thus, $\Pi_1 < \Pi_2 < \Pi_3 < \Pi_4$. The origin, $T=0, ~ M=0$, represents an isoperf corresponding to the {\em zero-shot} performance on $l$ when $\mathcal{M}$ is fine-tuned only on $P$.

The blue lines represent the isocosts. Considering the nature of the cost function defined, the isocost curves will be straight lines parallel to each other with slope $-\frac{c_t}{c_m}$. Like isoperfs, the cost of operation increases for the isocosts as we move towards right and top in the T-M diagram. 

\begin{define}
{\bf Least Cost Operating Point} on an isoperf refers to the (possibly multiple) point where the total cost of operation is lowest for a given performance.
\end{define}
 Under the assumption of smooth and convex isoperfs,\footnote{Isoperfs are convex for declining marginal rate of technical substitution.} %Refer to Appendix for details}
 the isocost corresponding to the least cost of operation will be a {\em tangent} to the isoperf, and the optimal allocation of the $T$ and $M$ is given by the \textit{point of tangency}. The isocosts shown in Fig.~\ref{fig:exp_path} correspond to the least cost curves for respective isoperfs, and the points of tangency are represented by the points E1, E2, etc.
 
 \begin{define}
{\bf Expansion path} is a path connecting the point of tangency of different isoperf and isocost curves, tracing out the cost minimizing combination of the data resources with increasing performance and costs. 
 \end{define}
 Expansion paths are important in determining resource allocations strategies. For instance, when a higher budget is available for dataset expansion in  a particular language, should one invest more in translation or in manually collected data? And how does this equation change in the long run, as the system moves towards higher performances?
 
Thus, isoperfs and isocosts when studied collectively can help determine the allocation of the amount of translation and manual data for a desirable performance value that minimizes the cost of operation.

\subsection{Selecting a Functional Form for $\pi$}
\label{sec:perf_func_form}
In production analysis, one of the difficult problems is to decide on the functional form of the production function that can on one hand accurately represent the input-output relationship, and on the other, is amenable to close-formed analysis~\cite{griffin1987selecting}. Clearly, a linear  production function would be an inappropriate choice for $\pi(T,M)$, as $T$ and $M$ are not perfect substitutes of each other. A popular choice in such case is the Cobb-Douglas performance function~\cite{cobb1928theory}, which is of the form $T^\alpha M^\beta$. However, the two datasets do not have multiplicative, but rather an additive effect. Therefore, we propose the following performance function:

\begin{equation}
    \Pi = \pi(T,M) = a_{zs} + a_{t}T^{\alpha_t} + a_{m}M^{\alpha_m}
\end{equation}
where $a_{zs}, a_t, a_m \geq 0$ and $0 \leq \alpha_t, \alpha_m \leq 1$. The positive coefficients of the input factors are motivated by assuming that under a reasonable translation and manual annotation quality, the addition of data from these sources should not hurt the zero-shot performance which is given by $a_{zs}$ (when $T=M=0$). Bounding the exponents below 1 ensures that the performance is not allowed to increase linearly with increasing data in one of these sources, as we always see diminishing returns with respect to data for any machine learning model. %The zero-shot performance can be obtained from this equation setting $T, M = 0$, giving $\Pi = a_{zs}$.

The commonly used training setups can be obtained as special cases of the above equation. The translate-train setup, can be obtained by setting $T = P$ and $M = 0$ in the equation, giving $\Pi_{TT} = a_{zs} + a_tP^{\alpha_t}$. Similarly, $\Pi_{FS} = a_{zs} + a_mk^{\alpha_m}$ gives the few-shot setup with $k$ examples. We denote this functional form as \pfone (Additive Model with Unequal Elasticities).

The sample efficiencies of the two sources of data can be derived by taking partial derivatives and is given by:

\begin{align}
    \psi(T) &= \frac{a_t\alpha_t}{T^{1 - \alpha_t}} \\
    \psi(M) &= \frac{a_m\alpha_m}{M^{1 - \alpha_m}}
\end{align}

The expression for tangency point can be derived by setting $dM/dT |_{\Pi=\Pi_c}$ to the slope of the isocost, $-c_t/c_m$, which gives the following equation for the {\em expansion path}.
\begin{equation}
    M = \left(\frac{c_t a_m \alpha_m}{c_m a_t \alpha_t}\right)^{\frac{1}{1 - \alpha_m}}T^\frac{1-\alpha_t}{1-\alpha_m}
    \label{eq:mt_ratio}
\end{equation}
Thus, $M/T$ (also called the labor-to-capital ratio) increases with performance if $\alpha_m > \alpha_t$, remains fixed when $\alpha_m=\alpha_t$, and decreases with performance when $\alpha_m < \alpha_t$. Similarly, the ratio of costs of acquiring manually created data to translated data, $Mc_m/Tc_t$ is proportional to $a_m M^{\alpha_m} / a_t T^{\alpha_t}$, which is the ratio of the contributions of the two datasets to the performance $\Pi$.

More often than not, actual production systems are too complex to be modeled accurately with simple functional forms. We expect a similar situation, where \pfone might be well suited for modeling and visualizing the trends. However, to obtain the actual operating cost and expansion path that are practically useful, one would need to model the behavior of the performance function more accurately. To this end, we also experiment with Gaussian Process Regression (GPR) for defining the performance function. As we shall see in the next section, GPR is able to fit the data more effectively, though we shall stick to \pfone as the two show identical trends and the latter also allows us to gain deeper insights and richer visualizations.

\subsection{Some Practical Considerations}
\label{sec:prac_conc}
\begin{define}
{\bf Cost Ratio}, defined as $c_{t/m} = \frac{c_t}{c_m}$, is the relative cheapness of the translation data, when compared to the cost of obtaining a manually created data point. 
\end{define}
We expect the cost ratio to be much smaller than 1. However, both translation and manual annotation costs vary according to the complexity (in case of translation, just the lengths of sentences) of the task at hand. $c_m$ might also vary with the choice of the target language $l$, while $c_t$ can be assumed to be uniform across the languages supported by the commercial MT systems like Google or Bing. In the experiments for our case study, we calculate the expansion paths for different values of $c_{t/m}$ to systematically study the nature of the trade-offs between the two sources of data.

\noindent
\textbf{Realizable region}: The forms of the performance function as well as cost function defined above do not place any constraint on the values that the input factors, i.e. $T$ and $M$, can take, which means that the amount of data can be increased indefinitely in order to improve the performance. However, we are aware that the amount of translated data is upper bounded by the amount of pivot data available, i.e. $T \leq P$. While this constraint can be explicitly worked out into the equations (by replacing $T$ with $\min(T, P)$), we stick to the original forms to preserve the smoothness of \pfone. Instead, we define a realizable region $\mathcal{R} : T \leq P$, and if a tangency point lies outside $\mathcal{R}$ we explicitly search for the minimum cost point on the part of the isoperf curve that lies in the realizable region. Note that, in such cases the isocost curves corresponding to the minimum cost point will no longer be tangents to the corresponding isoperfs, and will usually lie at the boundary between the realizable and non-realizable regions.
%Experiments
% 3. Experiments (1.5(including all diagrams))
%     3.1 Dataset and TyDIQA : Why TyDIQA
%     3.2 Training Test Setup
%     3.3 Estimating the Perf Function Parameters (GPR and Isoquant fiting)
%     3.4 Results:
%         - GPR and the Perf Function on the same graph, goodness of the fit, footnote: Other functional forms were also tried)
%         - Isocosts and Isoperfs plots : 4 curves. 1. Swahili with 0.1 Cr 2. Sw wth 0.05 Cr 3. Te with Pivot size 3696 and 4. Te with Pivot size 2000. Notes: 1. 1 million points graph, showing the factors are not perfectly subtitutable 2. At fixed performance increments, unequal increases in cost (Law of diminishing return). 3. Performance vs Least Cost (Assuming some CR and cost of translation), three curves, only translation, only labelled and mixed. 4. Amount of data required for different pivot sizes  (5. Plot of alpha and beta and a and b for different languages)

%Todo: Change notation of cost functions

\section{Case-Study on TyDiQA-GoldP}
In order to understand the efficacy of the proposed framework, we conduct a case-study on a popular multilingual Question Answering task (cf. $\mathfrak{T}$) using TyDiQA-GoldP~\cite{clark-etal-2020-tydi} dataset and consider mBERT as the MMLM $\mathcal{M}$. In the following subsections, we provide the details of the task and training setup for generating the performance $\Pi$ for different combination of the input factors, the procedure for estimating the parameters of the performance functions, and the findings.

\subsection{Task and Dataset}
We consider the Minimum Answer Span Task from the {Typologically Diverse Question Answering} benchmark or TyDiQA-GoldP for conducting the experiments. The choice of this particular dataset stems from two main properties of the benchmark. First, question-answering tasks are amenable to translation. Secondly,  TyDiQA-GoldP is comprised of manually labelled datasets for nine typologically diverse languages. This enables us to study the effect of different amounts of manually-created data $M$ on the performance of the MMLM. The amount of $M$ varies significantly from language to language with 1.6k examples for Korean to 15k examples in Arabic. 3.7k examples are available for English which we shall consider as the pivot language $p$ in all the experiments. We use Azure Translator\footnote{\url{https://www.microsoft.com/en-us/translator/business/translator-api/}} to obtain the translated data $T$ in eight target languages. The answer span alignment between English and the translated languages are obtained based on the technique described in~\citet{huetalXTREME}. We measure the performance $\Pi$ as the average F1-score between the predicted and actual answer-spans for the test examples. 
%For a detailed account of the task, languages supported and dataset statistics we point the reader to the original TyDiQA paper \cite{clark-etal-2020-tydi}.

\subsection{Fine-tuning Setup}
\label{sec:ft_setup}
We fine-tune mBERT on the TyDiQA-GoldP dataset with different values of the input factors, $T$ and $M$, for each target language, along with the amount of English pivot data, $P$.
%To fit the performance functions, we first fine-tune mBERT on the TyDiQA-GP dataset with different values of the input factors i.e. the amount of translated data and the amount of manually collected data, as well as consider the different input conditions like the amount of pivot data and the target language for which the performance is evaluated. 
Different values of $T$ are chosen by translating 0\%, 10\%, 40\% , 70\% or 100\% of the English pivot data. Eleven different values in the range $[0, |\mathcal{D}_l|]$ ($\mathcal{D}_l$ is the size of the available training data in $l$) and seven values between 0 and 3.7k are selected 
%(cf. Appendix) 
for $M$ and $P$, respectively. Considering eight different target languages, this results in 3080 different fine-tuning configurations. 
%As we will see in the results section, our framework provides a succinct way to summarize and analyze the results for these large number of settings.
 In each configuration, we use 3 different random seeds and train for 5 epochs with a learning rate of 2e-5 and a batch size of 32. The models are also jointly trained\footnote{We empirically observed that joint training performs better than curriculum learning ($P\rightarrow T \rightarrow M$)}. We use XTREME repository~\cite{huetalXTREME} and the Hugging Face Transformer Library~\cite{wolf-etal-2020-transformers} to conduct all our experiments.
%We follow curriculum based fine-tuning procedure i.e. sequentially train first on Pivot Data and then on Translated or Manual Data~\cite{lauscher-etal-2020-zero}.
%However, we did observe joint training to often performs on par and sometimes even better in our initial experiments so we proceed with that.

\subsection{Parameter Estimation of the Performance Function}
Upon estimating the performance values for the various fine-tuning configurations, we formulate the parameter estimation for the performance functions $\pi$ as a regression task, with $T$ and $M$ as inputs and $\Pi$ as the output. we use a Non Linear Least Squares algorithm~\cite{levenberg1944method} to fit the \pfone functional form (cf. Equation (5)), while specifying the bounds on the function parameters. For GPR, we use an RBF Kernel added with a White Kernel to model the noise in the observations, and the kernel parameters are optimized using L-BFGS-B optimization algorithm \cite{byrd1995limited} with 10 restarts. Note that, we fit different performance functions for each combination of $l$ and $P$. Additionally, we also conducted several experiments with other functional forms including Cobb-Douglas, linear, log-linear and polynomial functions ( > 1 degree) which either showed higher margins of error or over-fitting. 
%For all of these different alternatives we observe significantly worse fits with high margins of errors or in some cases like for cubic polynomials we observe a severe case of over-fitting (adding regularization often reduces the function to be linear).

\subsection{Results}

First, we evaluate how well the two proposed performance functions are able to predict the performance for different fine-tuning configurations. For this, we split the 3080 different training configurations into training (80\%) and test (20\%) sets.
The test root mean squared error (RMSE) and coefficient of determination ($r^2$) values for \pfone and GPR were found to be 5.84, 0.90 and 2.43, 0.98 respectively. Thus, both the models can fit the data reasonably well, though as expected, GPR provides a better fit. Check Appendix for more details. 

%We observe that both the models can fit the data reasonably well with the training and test $r^2 \geq 0.9$, though GPR obtains much lower errors. We also show the training errors corresponding to different fine-tuning setups including the two special cases of few-shot ($\Pi_{FS}$) and translate-train ($\Pi_{TT}$) in Table~\ref{tab:fitness_results}.
%Figure~\ref{fig:data_fit} presents a typical example of the \pfone and GPR based performance function estimates vis-\'{a}-vis the actual performance values for different $M$ and a fixed $T$.
%like few-shot , translate-train etc (which are just the special cases of our setup) which indicates that our models can accurately fit different regions of the performance landscape.
%The point is again illustrated in Figure\ref{fig:data_fit} which compares the predictions of \pfone and GPR with the actual F1-scores for different values of the amount of manual data (i.e. $M$), keeping $T$, $P$, and $p$ as fixed. 

\begin{table}[]
    \centering
    \begin{tabular}{lrrrr}
    \toprule
    $l$ &       $a_{t}$ &     $\alpha_t$ &      $a_m$ &  $\alpha_m$ \\
    \midrule
    \multicolumn{5}{c}{$P = 3696$}\\
    \midrule
     ar &  3.7e-01 &   1.9e-07 &  2.0e+00 &   2.2e-01 \\
    bn &  5.8e-04 &   6.9e-01 &  2.3e+00 &   3.0e-01 \\
     fi &  7.4e-02 &   3.9e-01 &  1.2e+00 &   3.0e-01 \\
     id &  2.5e-13 &   2.5e-01 &  1.2e+00 &   2.9e-01 \\
     ko &  2.6e-15 &   2.1e-03 &  1.5e+00 &   2.6e-01 \\
     ru &  7.8e-13 &   5.6e-01 &  7.1e-01 &   3.5e-01 \\
     sw &  5.2e-02 &   4.2e-01 &  1.1e+00 &   3.7e-01 \\
     te &  5.1e-19 &   2.5e-01 &  1.2e+01 &   1.5e-01 \\
    \midrule
    \multicolumn{5}{c}{$P = 2000$}\\
    \midrule
    ar &  1.7e-01 &   2.9e-01 &  2.9e+00 &   2.1e-01 \\
     bn &  9.9e-01 &   1.2e-01 &  1.9e+00 &   3.4e-01 \\
     fi &  9.4e-02 &   4.6e-01 &  1.6e+00 &   3.0e-01 \\
     id &  4.0e-01 &   1.2e-01 &  1.5e+00 &   3.0e-01 \\
     ko &  3.0e-13 &   4.1e-01 &  1.6e+00 &   2.8e-01 \\
     ru &  5.8e-03 &   6.5e-01 &  1.1e+00 &   3.4e-01 \\
     sw &  9.2e-02 &   4.3e-01 &  1.2e+00 &   3.7e-01 \\
     te &  1.6e-01 &   3.0e-01 &  1.2e+01 &   1.5e-01 \\
    \bottomrule
    \end{tabular}
    \caption{Values of \pfone performance function parameters for different languages.}
    \label{tab:params}
\end{table}

\begin{figure*}
    \centering
    \begin{subfigure}[t]{0.45\textwidth}
    \includegraphics[width=0.95\textwidth]{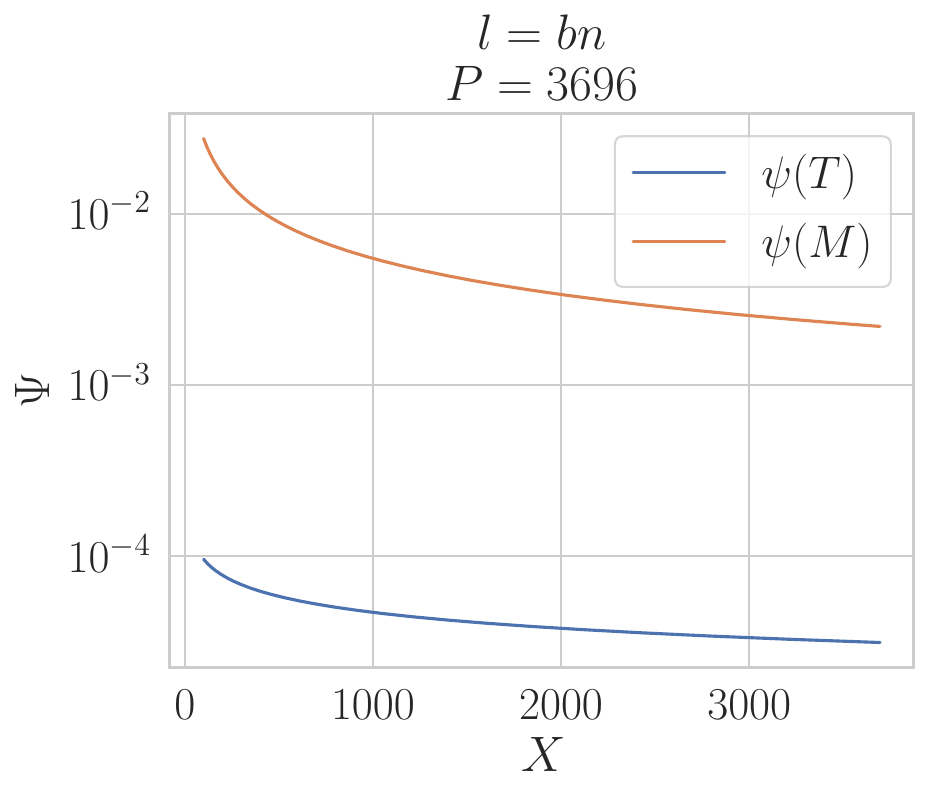}
    \caption{}
    \label{fig:sample_eff_bn_3600}
    \end{subfigure}
    \begin{subfigure}[t]{0.45\textwidth}
    \includegraphics[width=0.95\textwidth]{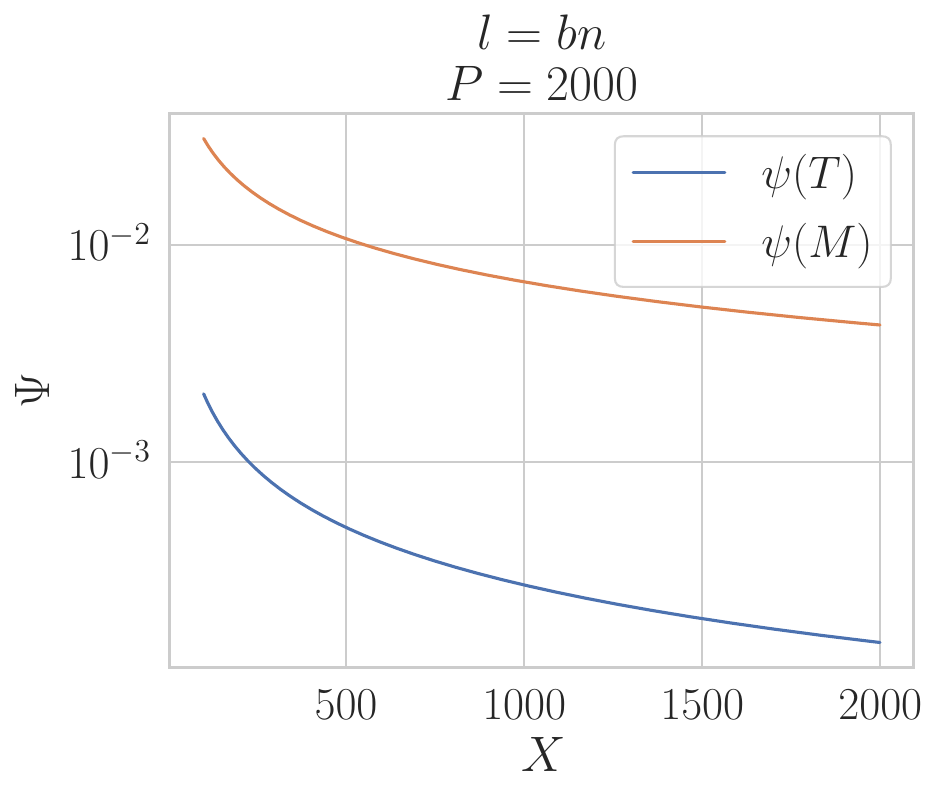}
    \caption{}
    \label{fig:sample_eff_bn_2000}
    \end{subfigure}
    \caption{Sample Efficiency Plots ($\Psi$ vs $X$) for $bn$ under different pivot sizes i.e. a) $P = 3696$, and b) $P = 2000$. Similar plots for other languages can be found in Figures \ref{fig:sample_eff_all_3696} and \ref{fig:sample_eff_all_2000} of Appendix.}
    \label{fig:sample_eff_bn}
\end{figure*}

\begin{figure*}
    \centering
    \begin{subfigure}[t]{0.45\textwidth}
    \includegraphics[width=0.95\textwidth]{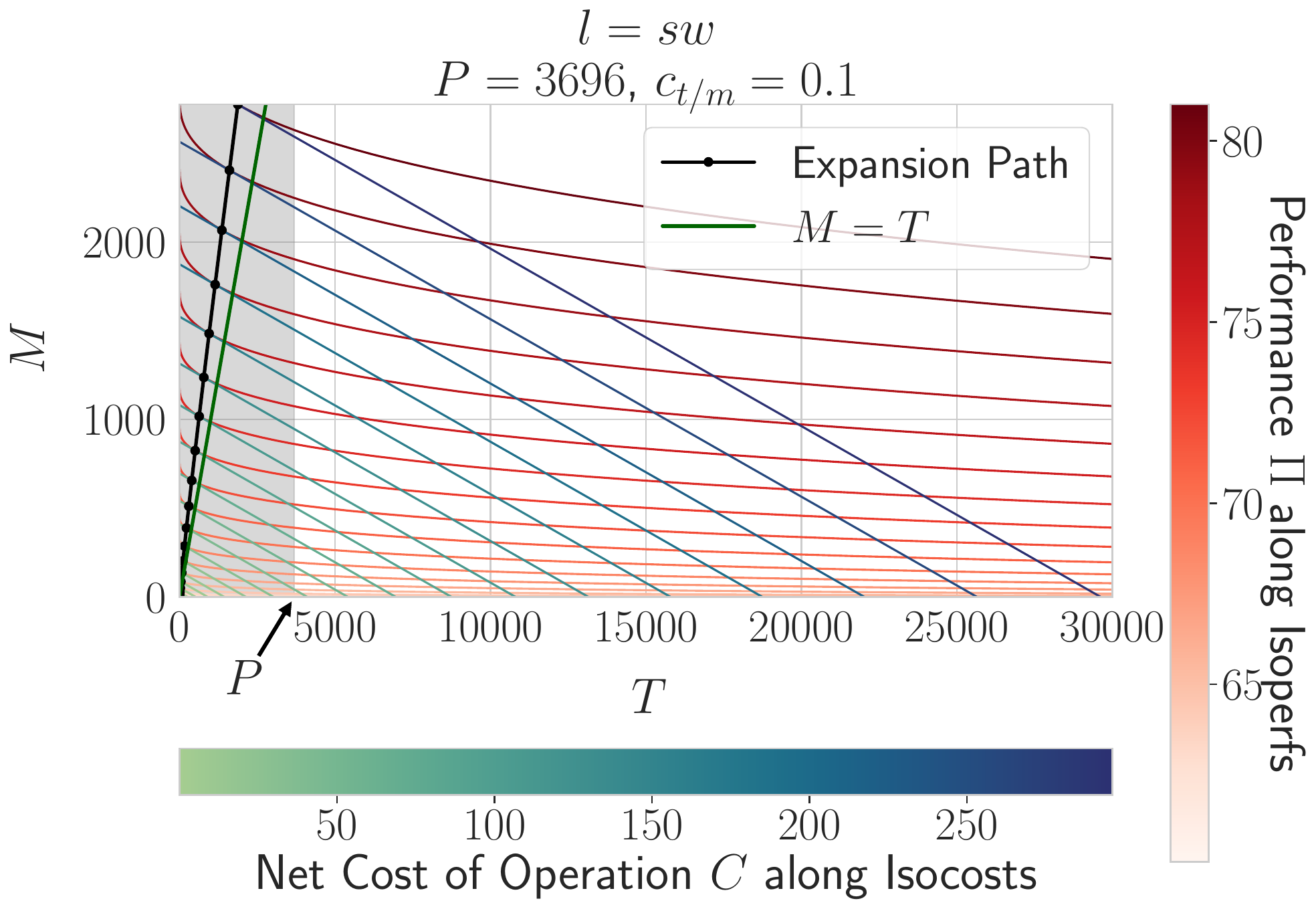}
    \caption{}
    \label{fig:sw_cr01}
    \end{subfigure}
    \begin{subfigure}[t]{0.45\textwidth}
    \includegraphics[width=0.95\textwidth]{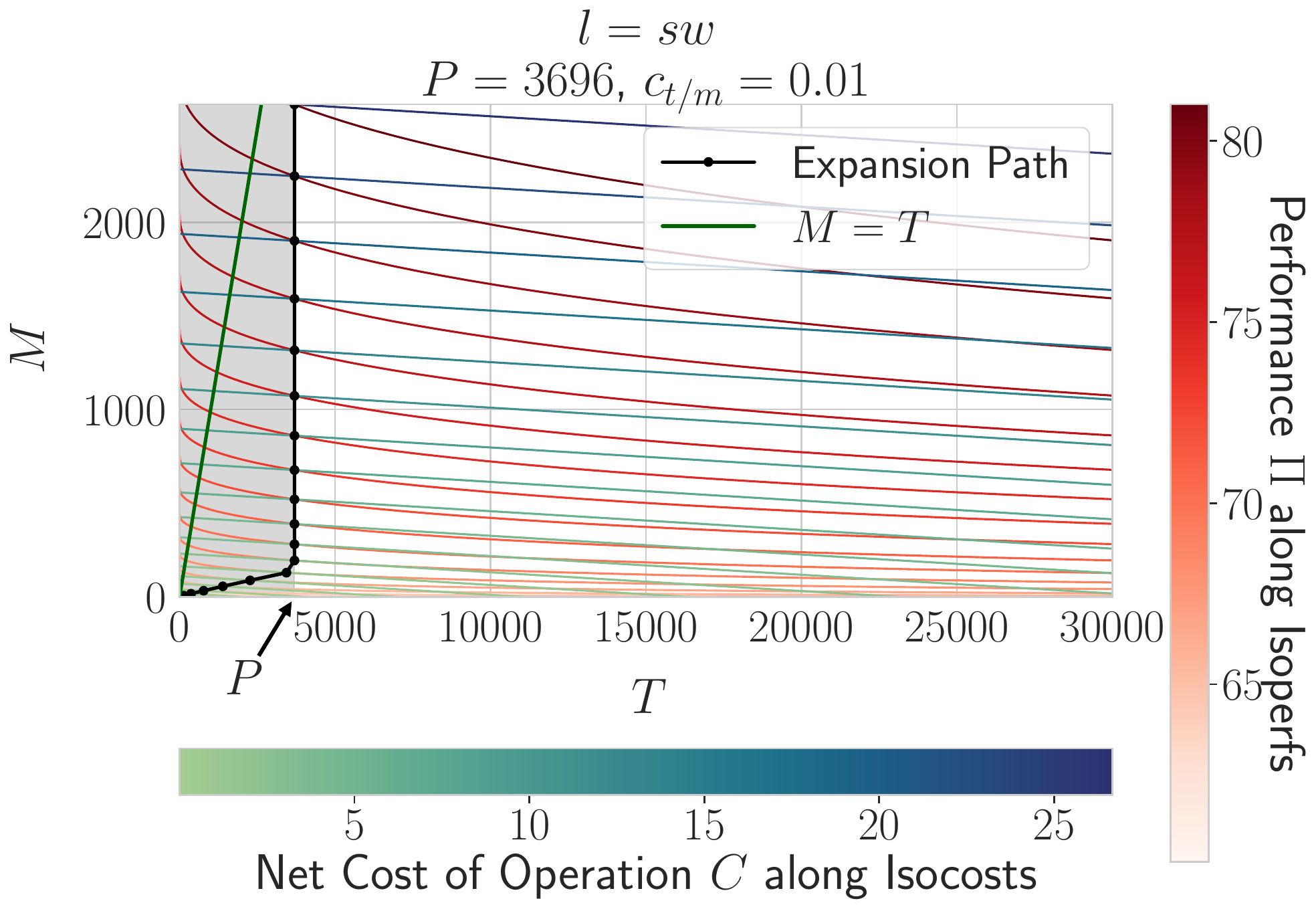}
    \caption{}
    \label{fig:sw_cr001}
    \end{subfigure}
    \begin{subfigure}[t]{0.45\textwidth}
    \includegraphics[width=0.95\textwidth]{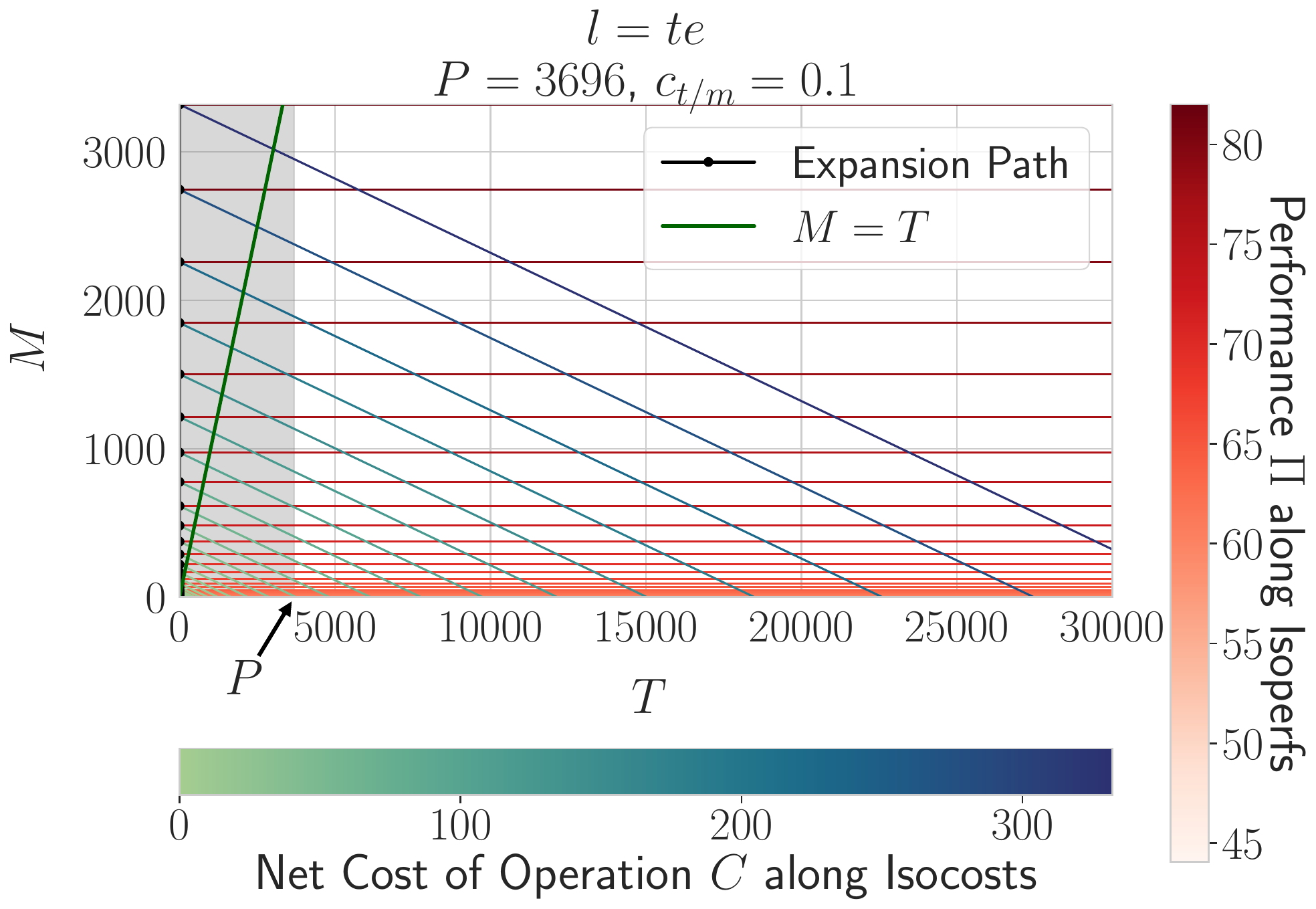}
    \caption{}
    \label{fig:te_ps3696}
    \end{subfigure}
    \begin{subfigure}[t]{0.45\textwidth}
    \includegraphics[width=0.95\textwidth]{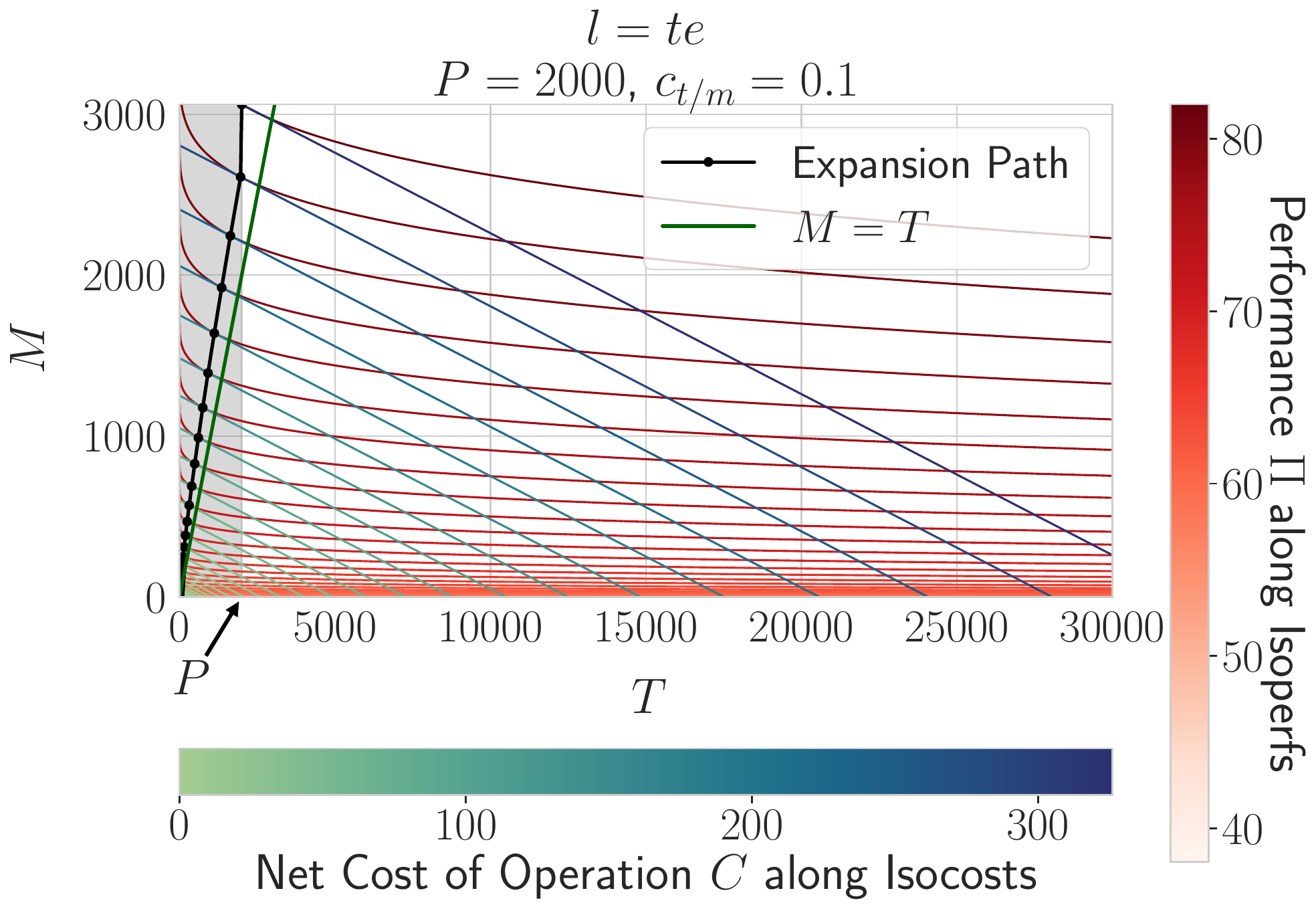}
    \caption{}
    \label{fig:te_ps2000}
    \end{subfigure}%
    \caption{M-T diagrams showing expansion paths obtained through \pfone for Swahili and Telugu for different values of $P$ and $c_{t/m}$. The shaded region represents ${\mathcal{R}}$ (cf. Sec.\ref{sec:prac_conc}).}
    \label{fig:exp_paths}
\end{figure*}

\noindent
\textbf{Sample Efficiency}: Table \ref{tab:params} shows the estimated values of the \pfone parameters for different languages and pivot sizes.  For all the languages, $a_m$ is greater than $a_t$ by at least an order of magnitude, meaning that the manually collected data ends up having a significantly higher contribution towards the model's performance. This can also be neatly visualized by the $\Psi - X$ diagrams in Figure \ref{fig:sample_eff_bn}, where for Bengali, the sample efficiencies of manual data is around two orders of magnitude higher.  For $P = 2000$, we see comparatively higher values of $a_t$ (though still $< a_m$) (Compare Figure \ref{fig:sample_eff_bn_3600} with Figure \ref{fig:sample_eff_bn_2000} ).  This indicates that the machine-translated data might be more beneficial when there is a paucity of training data available in the pivot language, and thus a lower zero-shot performance to begin with. However, the sample efficiency of manual data is still substantially higher.

\textbf{Expansion Paths}: For $P = 3696$, Arabic, Indonesian and Korean has $\alpha_m > \alpha_t$ and therefore, the corresponding expansion curves (Eqn~\ref{eq:mt_ratio}) will have an increasing $M/T$ ratio with increasing $\Pi$. %implying that the amount of $M$ with respect to the translations will continue to rise as we seek higher performance values at optimum costs. 
On the other hand for Swahili, Telugu and Finnish, $\alpha_m < \alpha_t$, and hence the expansion curves will bend towards the x-axis in the T-M diagram, indicating a declining $M/T$ ratio. In such cases, as we continue to increase the performance at the minimum cost, the optimum strategy would be to collect higher and higher amount of translation data as compared to manually labelled data.

However, notice that the $\alpha_m$ and $\alpha_t$ are close to each other for majority of the cases resulting in nearly linear expansion paths, a situation that is often encountered in economics whenever the production function is {\em homogenous}. We did not start with a homogenity assumption on $\pi(M,T)$; rather, the estimated parameters indicate so. This has two interesting implications: 1) $M/T$ remains nearly uniform at the different levels of performance; 2) the slope of the expansion path is approximately $(\frac{c_ta_m}{c_ma_t})^{\frac{1}{1-\alpha_m}}$ (by setting $\alpha_m=\alpha_t$ in Eqn~\ref{eq:mt_ratio}), meaning if the cost ratio $\frac{c_t}{c_m}$ is greater than $\frac{a_t}{a_m}$, the optimal strategy would be to collect more manually labelled data (since $\frac{1}{1-\alpha_m}$ > 1 by definition) and vice-versa. Thus, by just looking at the value of these parameters we can gain key insights about the optimal data allocation strategies. 

These strategic insights can also be clearly visualized through the isoperf, isocost and expansion path curves on the T-M diagrams, as shown in Fig. \ref{fig:exp_paths}. Due to paucity of space, we show the diagrams for two languages -- Swahili (sw) and Telugu (te) -- with two different cost ratios for the former (Fig.~\ref{fig:sw_cr01} and \ref{fig:sw_cr001}), and two different pivot sizes for the latter (Fig.~\ref{fig:te_ps3696} and \ref{fig:te_ps2000}). Refer appendix (\ref{fig:p3kc01},\ref{fig:p2kc01}, \ref{fig:p3kc001}, \ref{fig:p2kc001}) for rest.

For $c_{t/m} = 0.1$, $l=$sw (Fig.~\ref{fig:sw_cr01}), the expansion path follows a straight line roughly with a slope $(\frac{c_ta_m}{c_ma_t})^{\frac{1}{1 - \alpha_m}} = 3.2$.
This indicates that even though $M$ is 10 times more expensive than $T$, the optimal allocation policy is to still collect about thrice as much amount of $M$ as $T$. However, for $c_{t/m}=0.01$, which is less than $\frac{a_t}{a_m}$, 
the slope of the expansion path drops to $\approx 0.08$, as demonstrated by the theoretical expansion path on the right side of the $M = T$ line in Fig.~\ref{fig:sw_cr001}. 
%In Fig.~\ref{fig:sw_cr001} we observe that the expansion path, for the isoperfs corresponding to F1 scores between 60 to 70, bends significantly away from the line $M = T$ towards its right side with an average slope of about $0.05$. 
This suggests that we can rely on collecting a higher amount of translation data to increase the performance in this case because the manually collected data is much more expensive. 
As we move to the performance values > 70, we reach at the boundary of the realizable region (marked by translucent gray rectangle), and can no longer keep on collecting more translation data to increase the performance as by definition $T \leq P$. Beyond this point, to increase the performance, collecting higher amounts of manual data becomes inevitable. 
%\textit{meaning even if translations can be obtained at significantly lower costs, it can only take you far enough}.

For Telugu, we study the effect of two different values of $P$ and keep $c_{t/m}$ fixed at 0.1. At $P= 3696$, the isoperfs are nearly parallel to x-axis with the expansion path lying along the line $T = 0$ (Fig.~\ref{fig:te_ps3696}), which is expected as $\frac{a_t}{a_m} \approx 0$ in this case (see Table~\ref{tab:params}). This particular expansion path indicates that data obtained by translating English examples into Telugu does not have any notable performance improvement, though demands additional cost. The optimal strategy in this case is to only collect manually annotated data. This is not entirely surprising; the translate-train setup in~\citet{huetalXTREME} also shows low F1-scores for Telugu than the zero-shot setup.\footnote{Note that this does not invalidate the assumption we made in section \ref{sec:perf_func_form}. \citet{huetalXTREME} fine-tuned their models only on translated data, while we do train them with English Data as well and observe similar performance as zero-shot for Telugu.}
Interestingly, when $P=2000$ (Fig.~\ref{fig:te_ps2000}), $T$ provides non-trivial performance gains. The expansion curve is bent slightly to the left of the $M = T$ line, similar to Fig.~\ref{fig:sw_cr01}. This trend of higher $a_t/a_m$ for lower $P$ is observable for all languages (Table~\ref{tab:params}).

\begin{figure}
    \centering
    \includegraphics[width=0.45\textwidth]{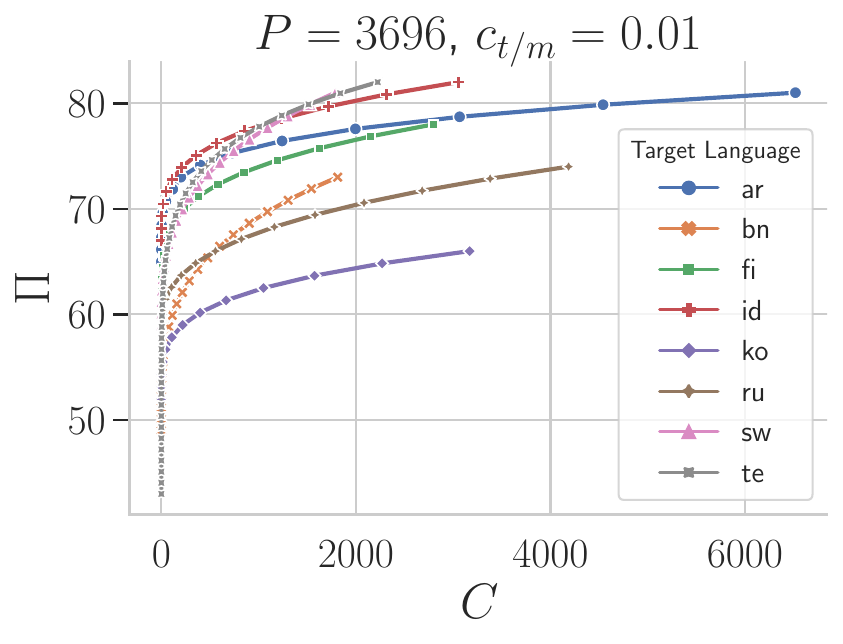}
    \caption{Performance vs the minimum costs for different languages. The performance function considered is \pfone. For $c=0.1$ case refer to Fig.~\ref{fig:cvspi_c01} in appendix.}
    \label{fig:cvspi}
\end{figure}

\noindent
\textbf{Performance and Cost Trade-off}:
Fig.~\ref{fig:cvspi} plots the cost vs the performance value traced out by the expansion paths for the 8 target languages. To calculate the total cost, we assume $c_t =  0.007$, which was estimated according to the standard translator Pricing offered by Azure\footnote{\url{https://azure.microsoft.com/en-us/pricing/details/cognitive-services/translator/}}, and consider $c_{t/m} = 0.01$. For all the languages, we observe a declining slope as we increase the value of $C$. Thus, it becomes increasingly more expensive to improve the performance of the models as we move to the higher values of $\Pi$ (\textit{law of diminishing returns}). 

% Further, for different languages the cost to performance ratios are also different. For instance, at a cost of 200\$, we can obtain about 17 points higher F1-score for Indonesian when compared with Korean. Similarly, a fixed F1-score of 70 can be obtained at 1/3rd of the cost compared to Russian. This indicates that different languages will require different level of economic investments to reach a desirable performance, which might be attributed to the factors like the varying complexity of the task in a particular language or the typological relatedness with the pivot language (which in this case is English). We should also point out that we have assumed uniform costs for manual annotations across the languages which might not be true in the practical settings as it might be more expensive to collect data in lower resource languages. However, this will just further add to disparity in the cost to performance ratio across the languages that we observe.

\begin{figure}
    \centering
    \includegraphics[width=0.45\textwidth]{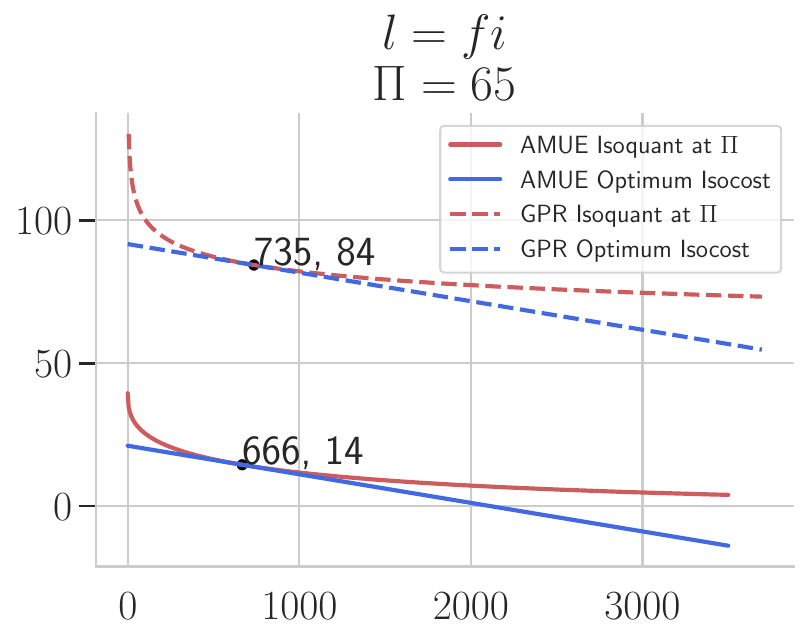}
    \caption{Comparing the Isoperfs and their corresponding optimum isocosts for \pfone and GPR production functions.}
    \label{fig:gprVamue}
\end{figure}

\noindent
\textbf{Comparing \pfone isoperfs with GPR isoperfs}: Figure \ref{fig:gprVamue} displays the isoperfs and the corresponding optimum isocosts obtained using \pfone and GPR based performance functions. As can be observed, both functions predict similar trends across their isoperfs; however, as expected, the curves are shifted due to different margin of errors for the two models.  

%Translations acting as substitutes when low amounts of data is available and might act as a form of data augmentation
%Discussion
% 4. Discussion (0.5 pages)
%     - Summary of findings
%     - Generalizability of the model and other applications beyond translation and labelling
%     - Generalizing it to multi-pivot and multi-target cases, multiple output production performance functions
%     - Multiple Technology. Fixed budget different models (mBERT, XLMR, monolingual bert), isoperfs would change
%     - Translation Quality
%     -[Optional] Understanding the dependence of tradeoffs with the linguistic properties of the languages. How generalizable these patterns are for different language and task, can we guess the expansion curves for the new tasks and languages (cite AAAI predictor paper)

\section{Discussion and Conclusion}

In this work, we have proposed a micro-economics inspired framework to study the performance and cost trade-offs between manually annotated and machine-translated data for training multilingual models, and demonstrated its efficacy through a case-study on the TyDiQA-GoldP dataset. The key findings from this case-study are:
\begin{enumerate*}
\item[1.] Some amount of manually collected data in a target language is crucial to attain optimal performance at minimum cost irrespective of how much cheaply MT data can be procured, as long as the cost is non-zero.
\item[2.] The ratio of manually collected and machine-translated data at least cost operating point remains nearly uniform at the different levels of performance % and is roughly proportional to the cost ratio $c_{t/m}$.
\item[3.] The usefulness of translated data is higher when the amount of pivot language data is less. 
\end{enumerate*}
There are several other insights that can be drawn from the T-M diagrams and other plots, which could not be presented here due to the paucity of space.

This work can be expanded in several ways. In the current work we considered a single-pivot and single-target case. Generalizing this to the case where the model is allowed to be trained on multiple pivot languages and then be evaluated on multiple targets is of considerable interest. This implies extension to multiple-output production functions with multiple ($> 2$) input factors.

Here, we have not considered the effect of \textit{multiple technology} on the isoperfs. For our problem, multiple technologies may correspond to the different MMLMs such as mBERT, XLMR and mT5, different MT systems, and even different training curricula. Identifying the optimal allocation policy considering the presence of such multiple technological alternatives would be an interesting exercise.
In particular, it will be interesting to explore the impact of translation quality on the trade-offs. %For instance, does the coefficient ratio $\frac{a_t}{a_m}$ improve for better translation quality and vice-versa.
%In similar way, the correlation of these parameters with the typological relatedness of the pivot and target language can also be studied. 
An important limitation of the current framework is that it presumes availability of certain amounts of M and T datasets such that the performance function can be estimated. However, in practice, one would like to understand the trade-offs before collecting the data. Recently, \citet{srinivasan2021predicting} showed that it is possible to predict the zero-shot and few-shot performance of MMLMs for different languages using linguistic properties and their representation in the pre-training corpus. Understanding if there exists a similar dependence of the performance trade-offs with the linguistic properties of different languages can help us generalize our framework to the new languages without the need for explicit data collection. 

Finally, we believe that performance function-based analysis can be applied to a multitude of three-way trade-offs among technology, cost and data that are commonly encountered in the NLP world. The economics of language data can be a new direction of study with important practical and theoretical applications.

\section*{Acknowledgements}
We would like to thank Shanu Kumar and the LITMUS team at Microsoft for their valuable inputs and feedback over the course of this work. We are also grateful to the anonymous reviewers for their constructive comments.

\bibliography{anthology,custom}
\bibliographystyle{acl_natbib}

\appendix

\label{sec:appendix}

\section{Appendix}

\subsection{Derivations}

Here we derive the expression for the curve traced by the expansion path as given in equation \ref{eq:mt_ratio}. As described in section \ref{sec:perf_func_form} \pfone performance function is given by:
\begin{equation*}
    \pi(T,M) = a_{zs} + a_{t}T^{\alpha_t} + a_{m}M^{\alpha_m}
\end{equation*}

Setting $\pi(T,M) = \Pi_c$ i.e. a constant value, we can obtain an analytic expression for the isoperf curves from this functional form, which is given by:
\begin{equation}
     M = \left(\frac{\Pi_c - a_{zs} - a_tT^{\alpha_t}}{a_m}\right)^{\frac{1}{\alpha_m}}
 \end{equation}
 
 Since the expansion path is the locus of the points of tangency between isoperf and isocost curves, we can compute the slope of the isoperf curve and set them equal to each other. The slope for isoperf curve can be computed as:
 
 \begin{align*}
    M^{\alpha_m} &= \left(\frac{\Pi_c - a_{zs} - a_tT^{\alpha_t}}{a_m}\right)\\
    \alpha_mM^{\alpha_m - 1} \diff{M}{T} &= -\frac{\alpha_ta_t}{a_m}T^{\alpha_t - 1}\\
    \diff{M}{T} &= -\frac{\alpha_ta_t}{\alpha_ma_m}\frac{T^{\alpha_t-1}}{M^{\alpha_m-1}}
\end{align*}

The slope of the isocost curve is simply $-\frac{c_t}{c_m}$, equating them we get:

 \begin{align*}
 \frac{c_t}{c_m} &= \frac{\alpha_ta_t}{\alpha_ma_m}\frac{T^{\alpha_t - 1}}{M^{\alpha_m - 1}}\\
 M^{\alpha_m-1} &= \frac{\alpha_ta_tc_m}{\alpha_ma_mc_t}T^{\alpha_t - 1}\\
M &= \left(\frac{c_t a_m \alpha_m}{c_m a_t \alpha_t}\right)^{\frac{1}{1 - \alpha_m}}T^\frac{1-\alpha_t}{1-\alpha_m}\\
 \end{align*}
 
 \subsection{Training Setup}

We typically run the fine-tuning experiments on NVIDIA-P100 GPUs with 16 GB of memory. A fine-tuning job with 3 random seeds typically takes 2 hours to run on the specified compute. Having access to 64 of such GPUs we ran multiple jobs in parallel. For fitting performance functions and doing analysis on expansion paths CPU only compute of Intel(R) Xeon(R) CPU E5-2690 was utilized.

We use mBERT configuration bert-base-multilingual-cased for fine-tuning, which supports 104 languages and has around 178 million parameters.

\subsection{Goodness of Fit}

\begin{figure}[htb]
    \centering
    \includegraphics[width=0.4\textwidth]{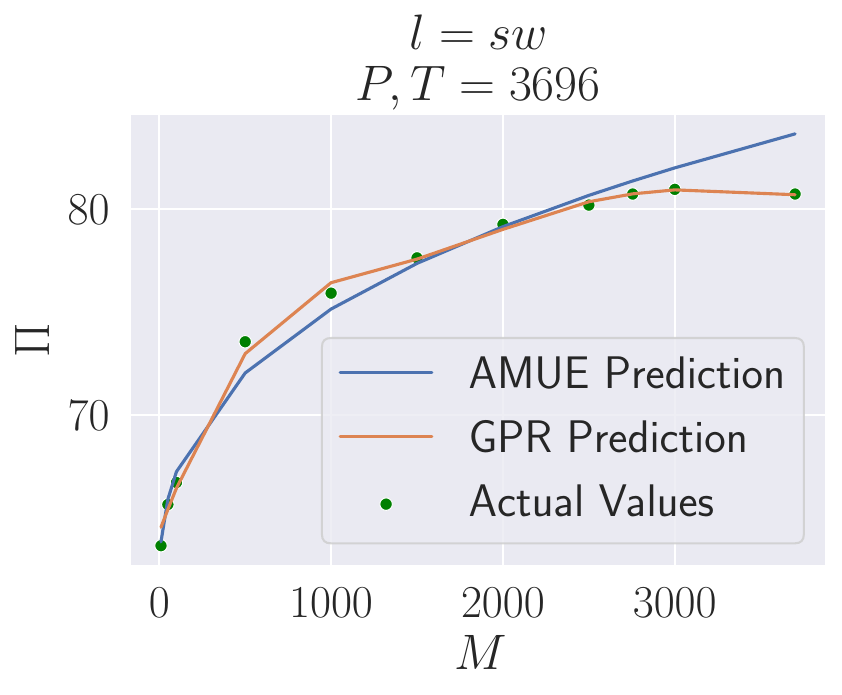}
    \caption{Performance function estimated by \pfone and GPR. $\Pi \equiv $ F1-score (scaled by 100). 
    %Both models can reasonably fit the performance data, however, GPR owing to its capacity to fit a wide variety of functions, does better.
    }
    \label{fig:data_fit}
\end{figure}

Table \ref{tab:fitness_results} shows the train and test RMSE and $r^2$ for GPR and \pfone. For training set we also compute the errors corresponding to different fine-tuning setups like translate-train , few-shot etc, which indicates that our models can accurately fit different regions of the performance landscape. The point is again illustrated in Figure\ref{fig:data_fit} which compares the predictions of \pfone and GPR with the actual F1-scores for different values of the amount of manual data (i.e. $M$), keeping $T$, $P$, and $p$ as fixed. 

\begin{table}[htb]
\centering
\scalebox{0.8}{
\begin{tabular}{ll cc cc}
    \toprule
    && \multicolumn{2}{c}{\thead{\pfone}} & \multicolumn{2}{c}{\thead{GPR}}\\
    \cmidrule(r){3-4}
    \cmidrule(r){5-6}
     \thead{Data Split} & \thead {Fine-tune \\ setup}  & \thead{RMSE $\downarrow$} & \thead{$r^2 \uparrow$} & \thead{RMSE $\downarrow$} & \thead{$r^2 \uparrow$}  \\
     \midrule
     \multirow{5}{2em}{\thead{Train}} & \thead{Zero-Shot} & \thead{4.19} & \thead{0.95} & \thead{4.43} & \thead{0.95}\\
     & \thead{Translate-Train} & \thead{5.10} & \thead{0.93} & \thead{3.68} & \thead{0.96}\\
     & \thead{Few-Shot} & \thead{5.75} & \thead{0.90} & \thead{1.63} & \thead{0.99}\\
     & \thead{Few-Shot \\ + Translate-train} & \thead{4.71} & \thead{0.93} & \thead{1.53} & \thead{0.99}\\
     & \thead{Overall} & \thead{5.04} & \thead{0.93} & \thead{1.86} & \thead{0.99}\\
     \midrule
     \thead{Test} & \thead{Overall} & \thead{5.84} & \thead{0.90} & \thead{2.43} & \thead{0.98}\\
     \bottomrule
\end{tabular}
}
\caption{RMSE and $r^2$ values for the two performance functions on training and test sets.}
\label{tab:fitness_results}
\end{table}

\begin{figure*}
    \centering
    \begin{subfigure}[t]{0.45\textwidth}
    \includegraphics[width=0.95\textwidth]{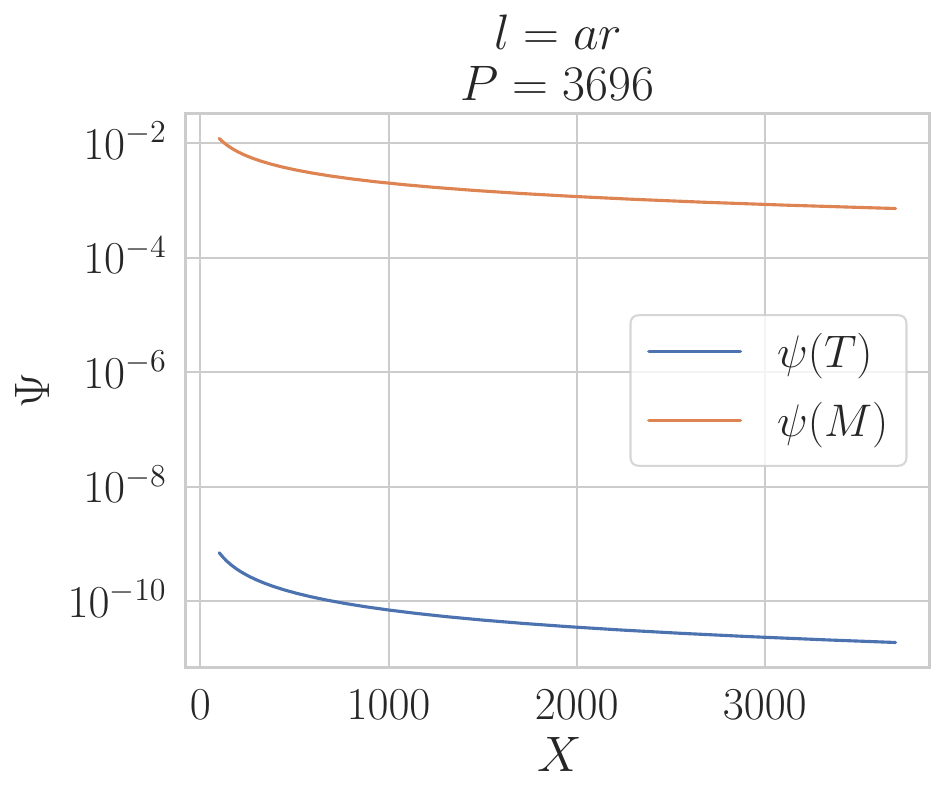}
    \caption{}
    \label{}
    \end{subfigure}%
    \begin{subfigure}[t]{0.45\textwidth}
    \includegraphics[width=0.95\textwidth]{figures/bn_3696.pdf}
    \caption{}
    \label{}
    \end{subfigure}
    \begin{subfigure}[t]{0.45\textwidth}
    \includegraphics[width=0.95\textwidth]{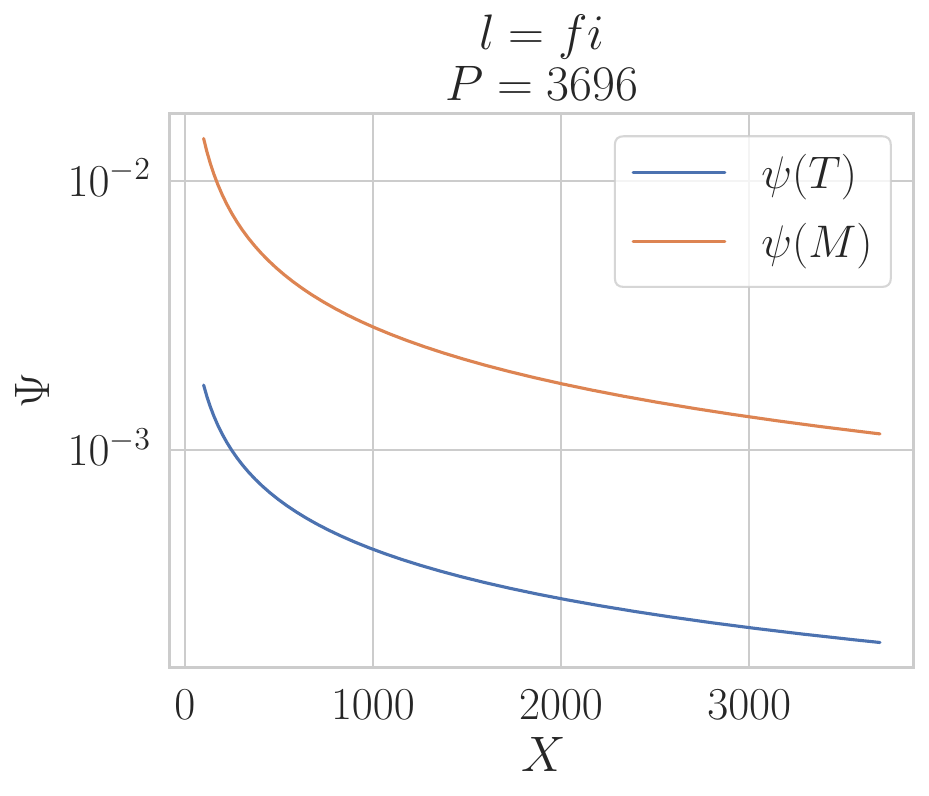}
    \caption{}
    \label{}
    \end{subfigure}%
    \begin{subfigure}[t]{0.45\textwidth}
    \includegraphics[width=0.95\textwidth]{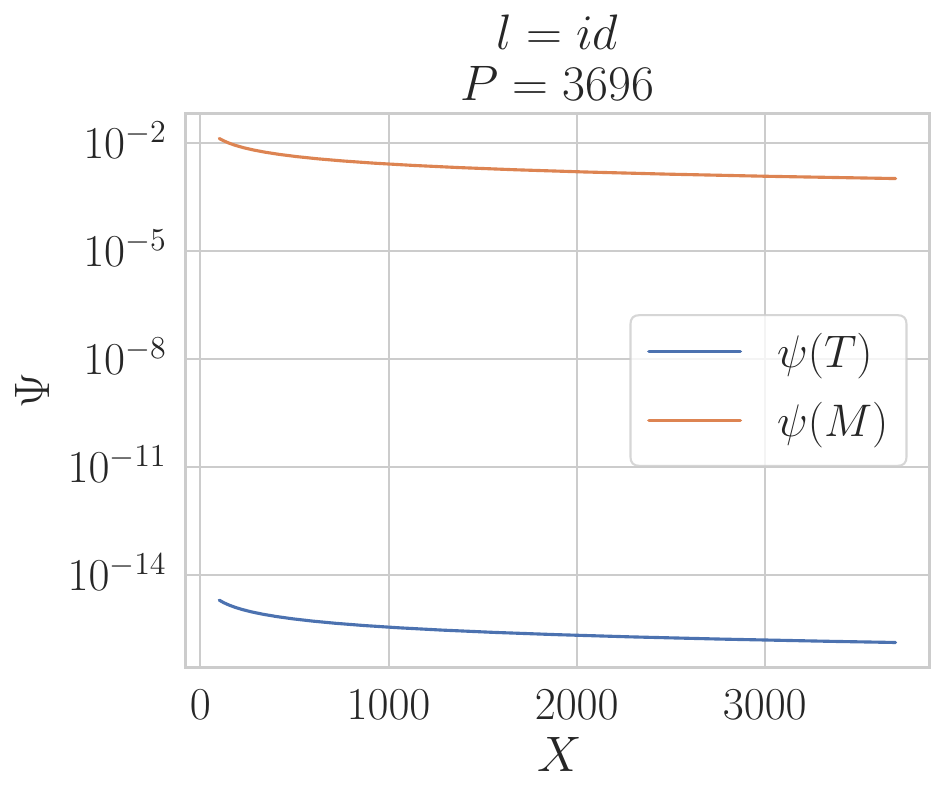}
    \caption{}
    \label{}
    \end{subfigure}
    \begin{subfigure}[t]{0.45\textwidth}
    \includegraphics[width=0.95\textwidth]{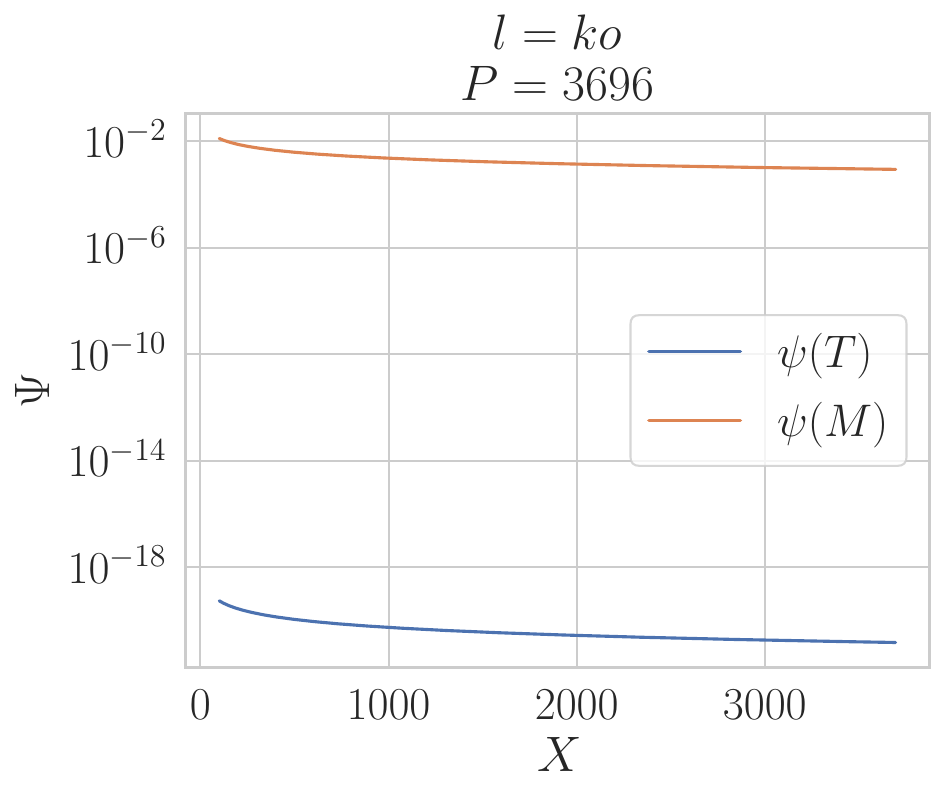}
    \caption{}
    \label{}
    \end{subfigure}%
    \begin{subfigure}[t]{0.45\textwidth}
    \includegraphics[width=0.95\textwidth]{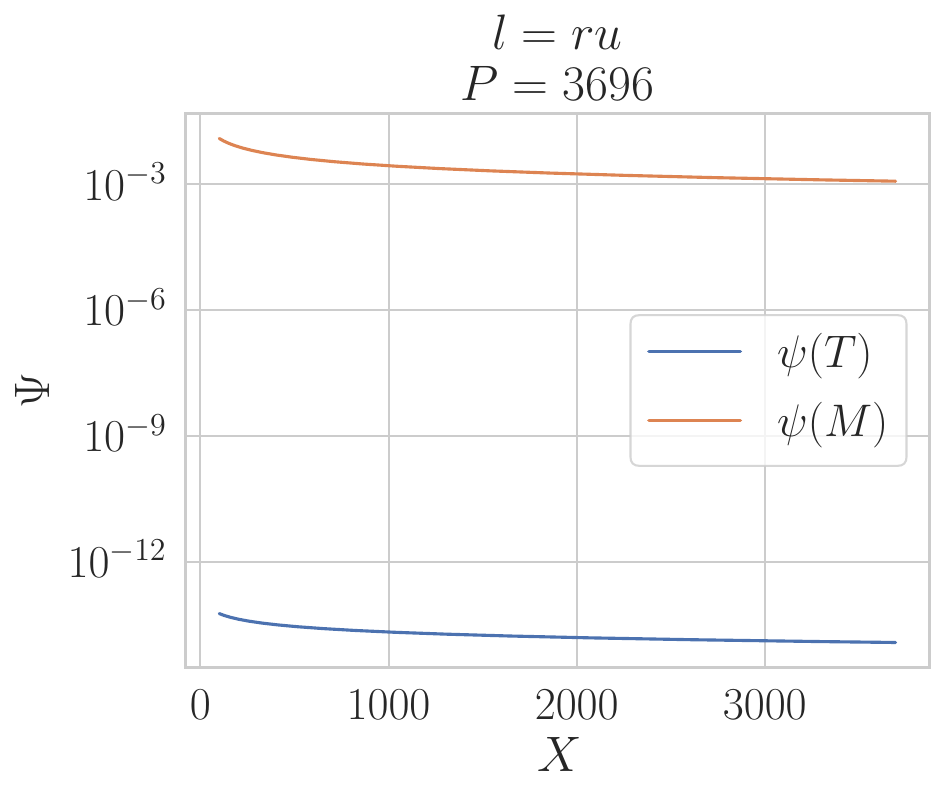}
    \caption{}
    \label{}
    \end{subfigure}
    \begin{subfigure}[t]{0.45\textwidth}
    \includegraphics[width=0.95\textwidth]{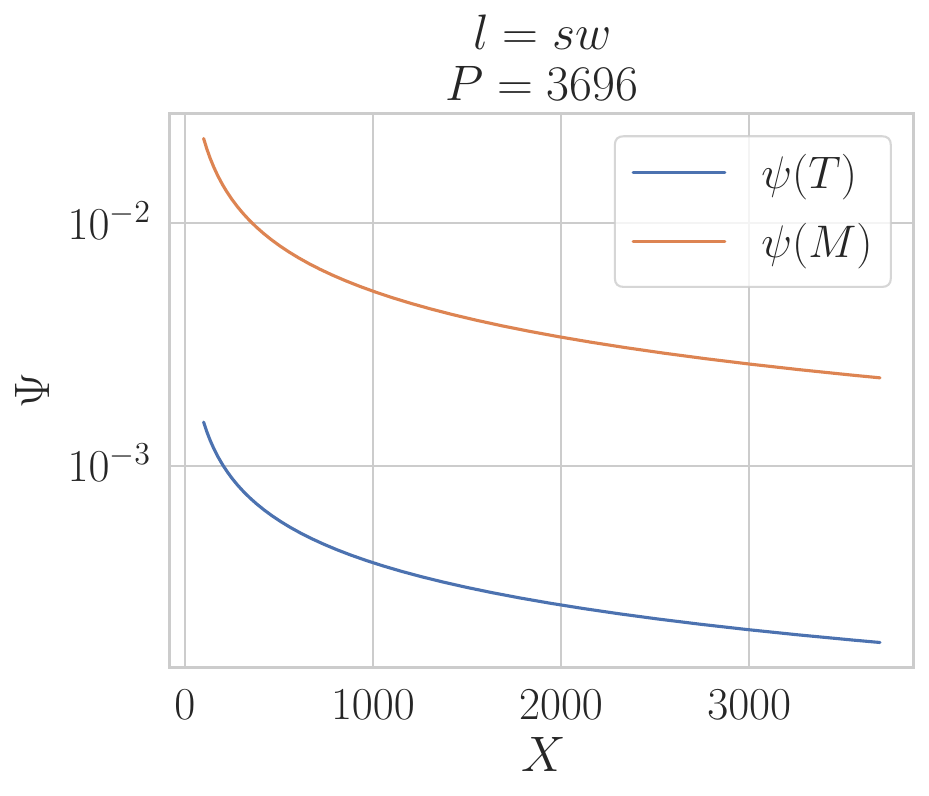}
    \caption{}
    \label{}
    \end{subfigure}%
    \begin{subfigure}[t]{0.45\textwidth}
    \includegraphics[width=0.95\textwidth]{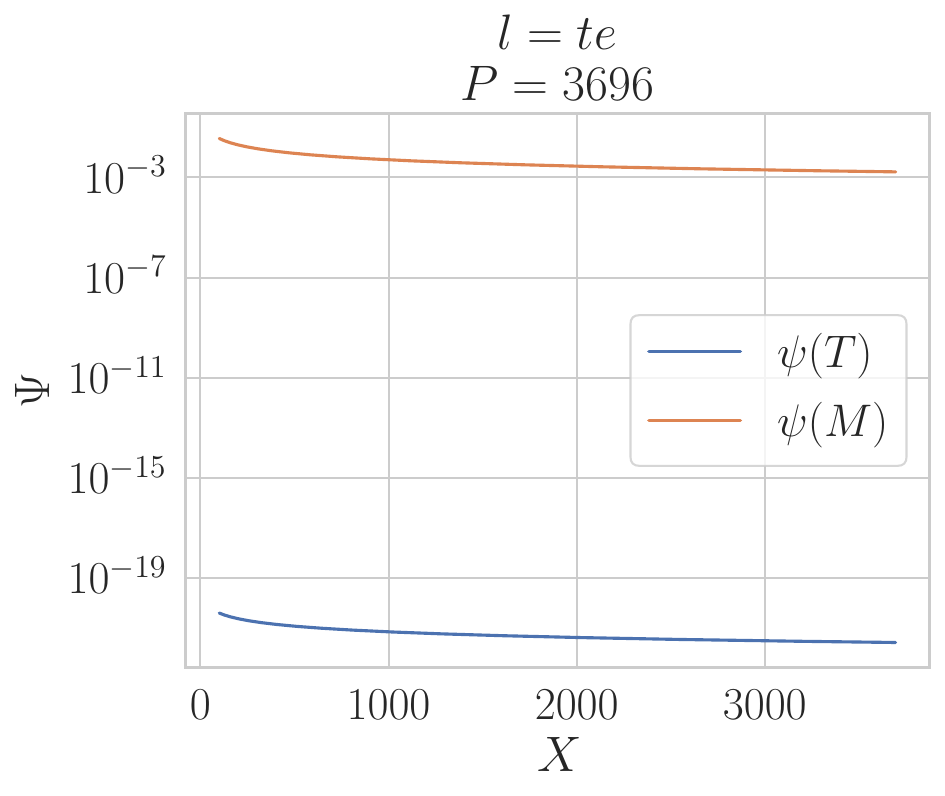}
    \caption{}
    \label{}
    \end{subfigure}
    \caption{Sample Efficiency ($\Psi-X$) plots for different languages for $P = 3696$}
    \label{fig:sample_eff_all_3696}
\end{figure*}

\begin{figure*}
    \centering
    \begin{subfigure}[t]{0.45\textwidth}
    \includegraphics[width=0.95\textwidth]{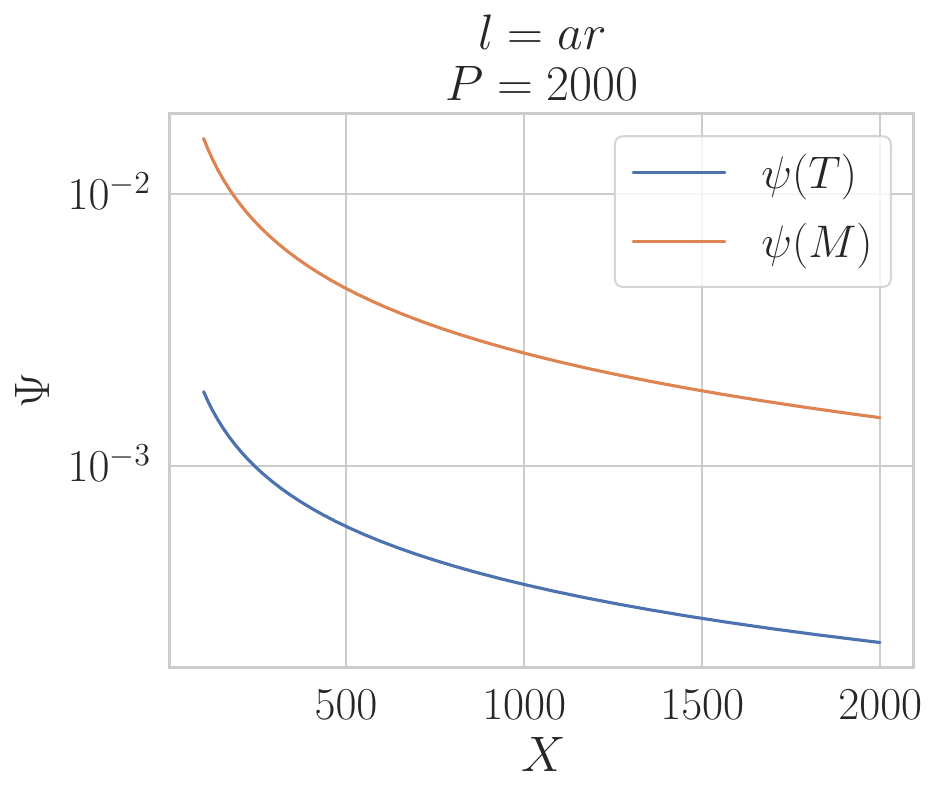}
    \caption{}
    \label{}
    \end{subfigure}%
    \begin{subfigure}[t]{0.45\textwidth}
    \includegraphics[width=0.95\textwidth]{figures/bn_2000.pdf}
    \caption{}
    \label{}
    \end{subfigure}
    \begin{subfigure}[t]{0.45\textwidth}
    \includegraphics[width=0.95\textwidth]{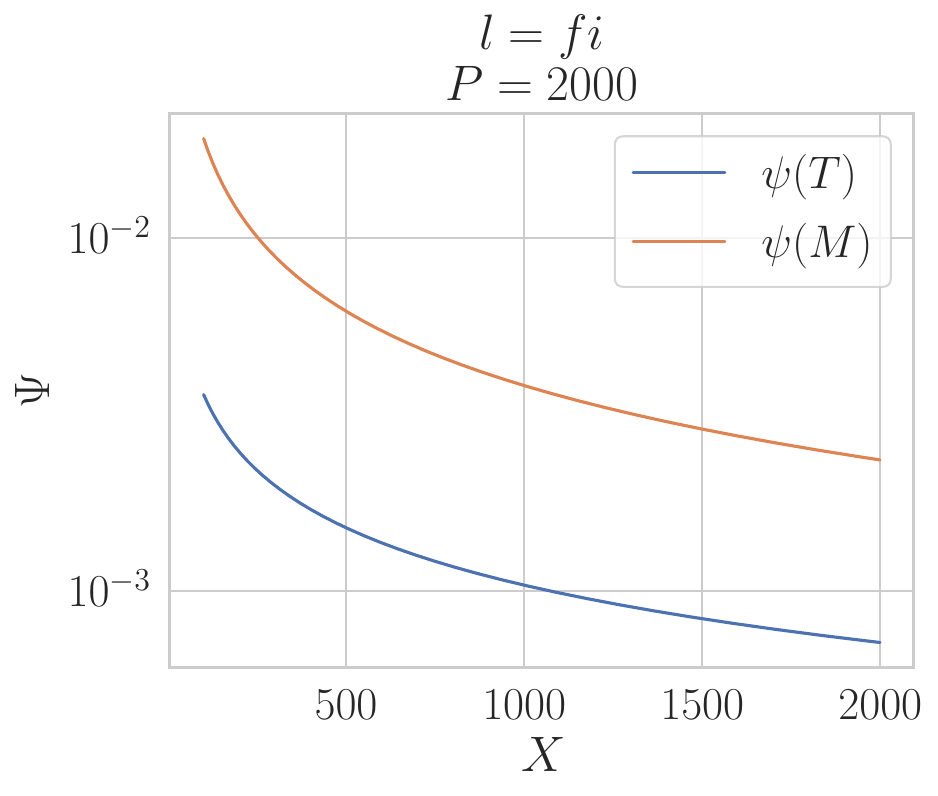}
    \caption{}
    \label{}
    \end{subfigure}%
    \begin{subfigure}[t]{0.45\textwidth}
    \includegraphics[width=0.95\textwidth]{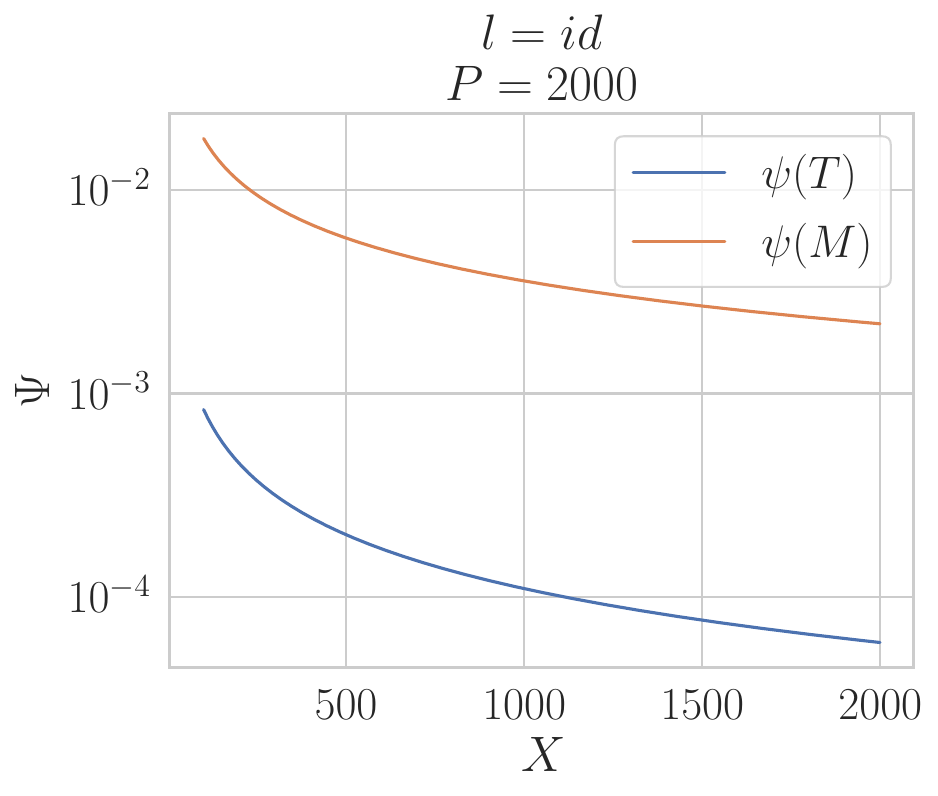}
    \caption{}
    \label{}
    \end{subfigure}
    \begin{subfigure}[t]{0.45\textwidth}
    \includegraphics[width=0.95\textwidth]{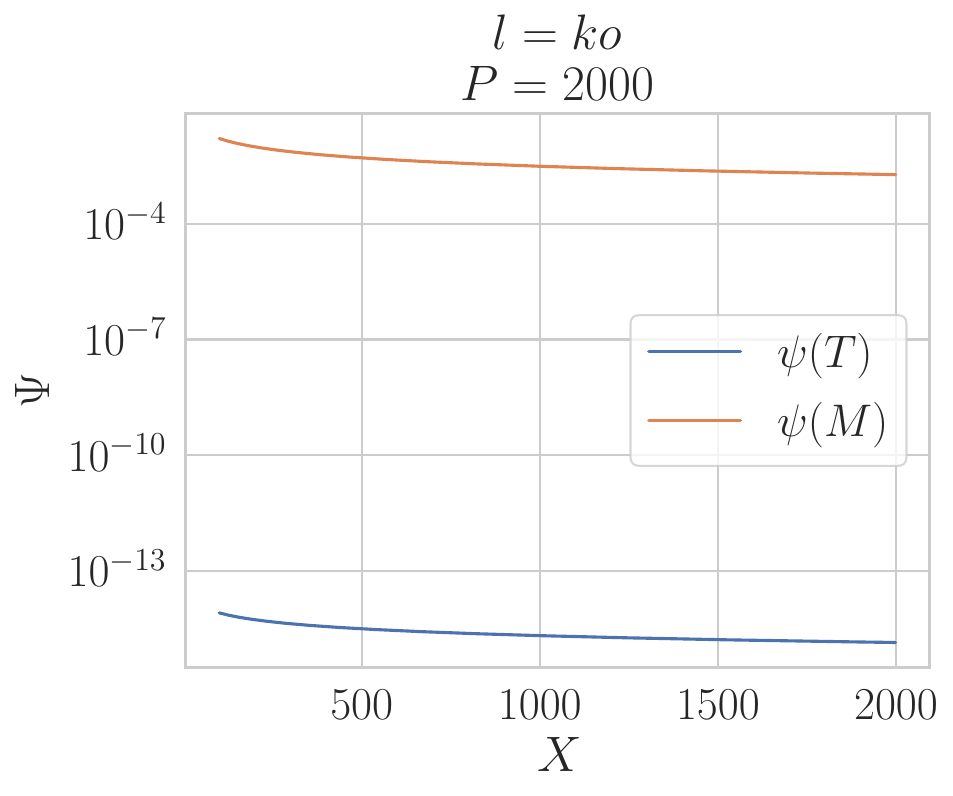}
    \caption{}
    \label{}
    \end{subfigure}%
    \begin{subfigure}[t]{0.45\textwidth}
    \includegraphics[width=0.95\textwidth]{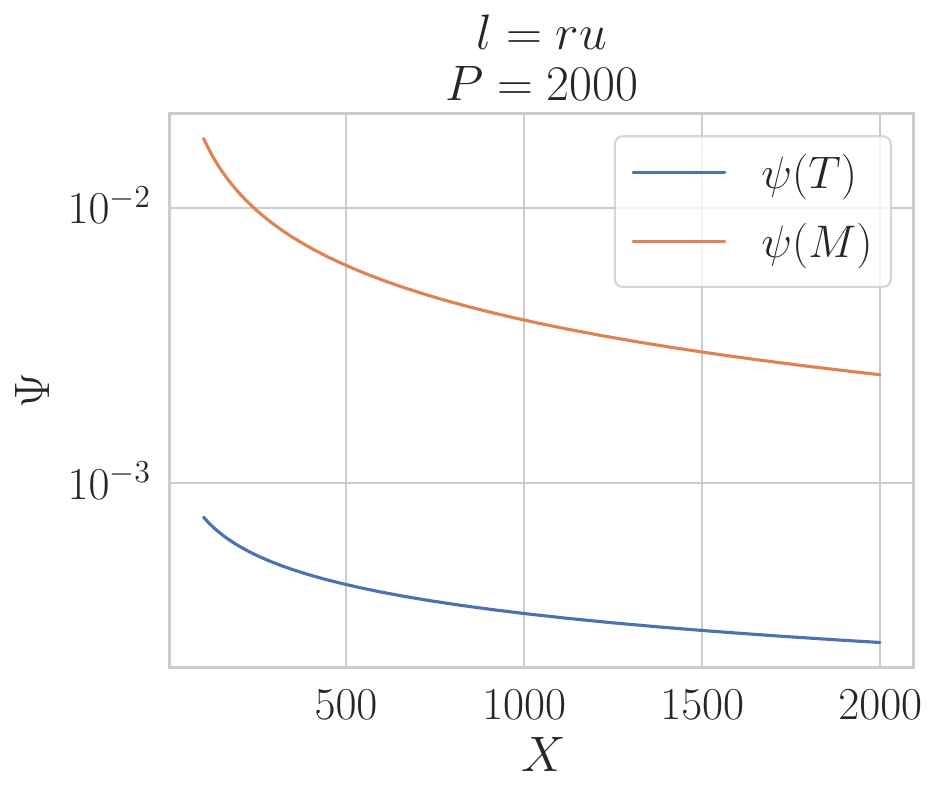}
    \caption{}
    \label{}
    \end{subfigure}
    \begin{subfigure}[t]{0.45\textwidth}
    \includegraphics[width=0.95\textwidth]{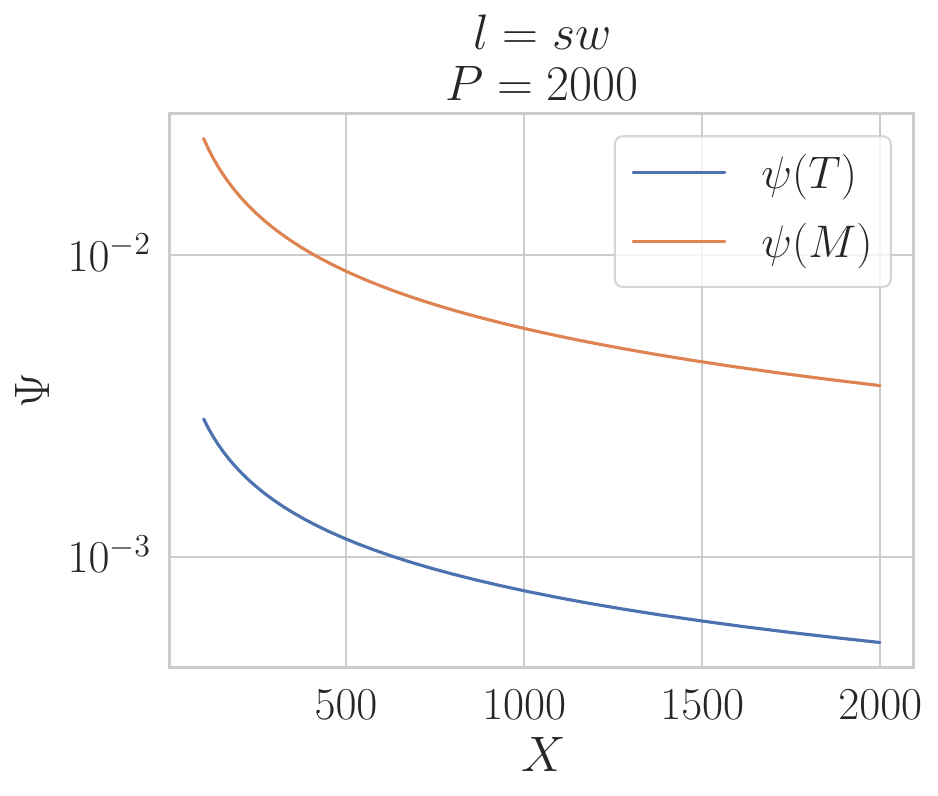}
    \caption{}
    \label{}
    \end{subfigure}%
    \begin{subfigure}[t]{0.45\textwidth}
    \includegraphics[width=0.95\textwidth]{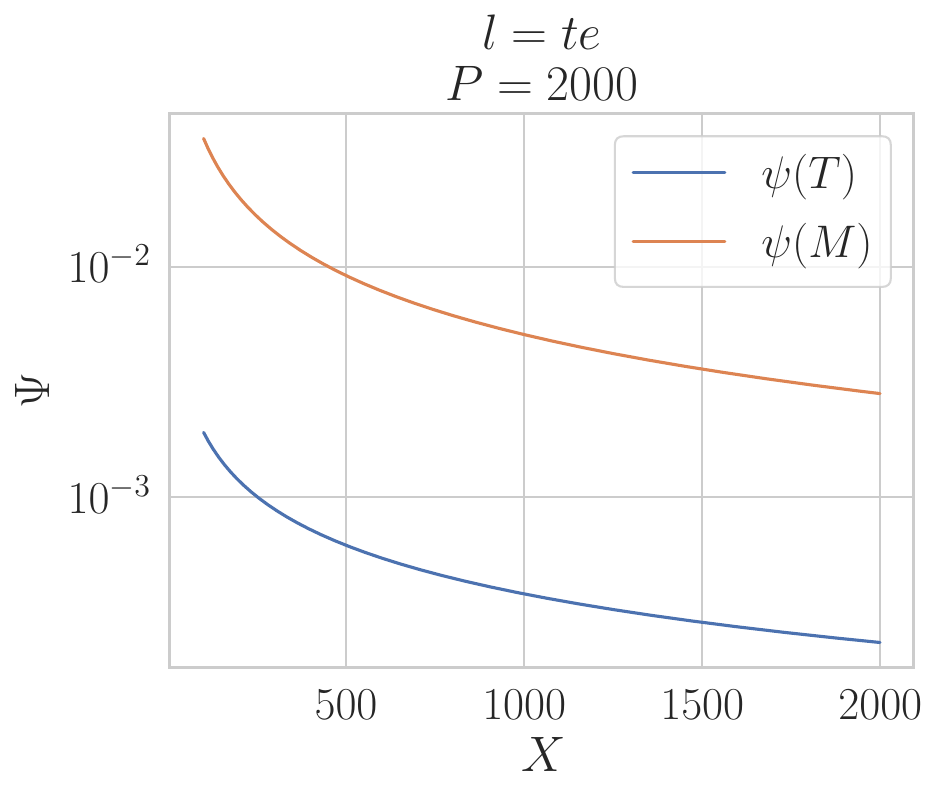}
    \caption{}
    \label{}
    \end{subfigure}
    \caption{Sample Efficiency ($\Psi-X$) plots for different languages for $P = 2000$}
    \label{fig:sample_eff_all_2000}
\end{figure*}

\begin{figure*}
    \centering
    \begin{subfigure}[t]{0.45\textwidth}
    \includegraphics[width=0.95\textwidth]{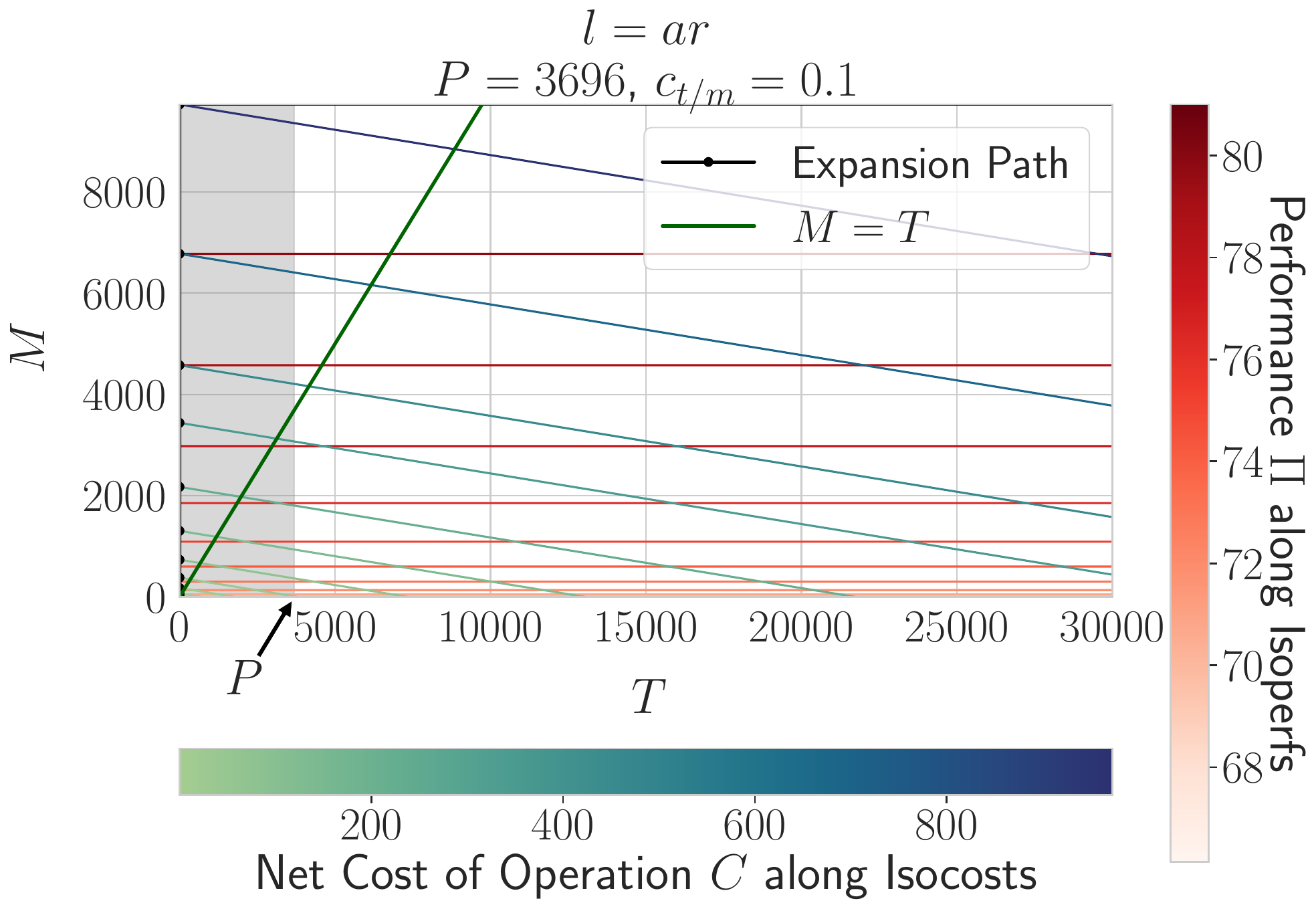}
    \caption{}
    \label{}
    \end{subfigure}%
    \begin{subfigure}[t]{0.45\textwidth}
    \includegraphics[width=0.95\textwidth]{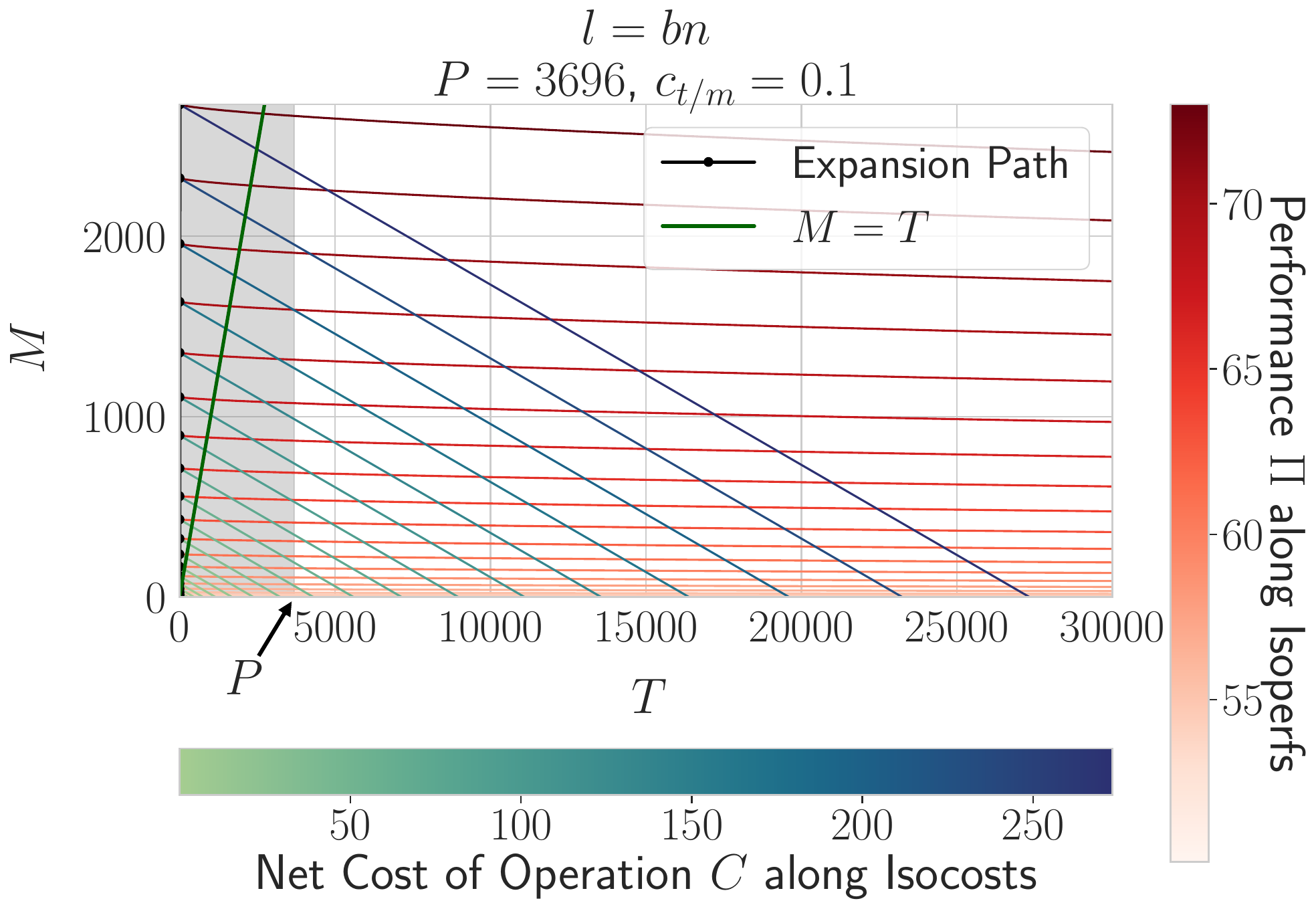}
    \caption{}
    \label{}
    \end{subfigure}
    \begin{subfigure}[t]{0.45\textwidth}
    \includegraphics[width=0.95\textwidth]{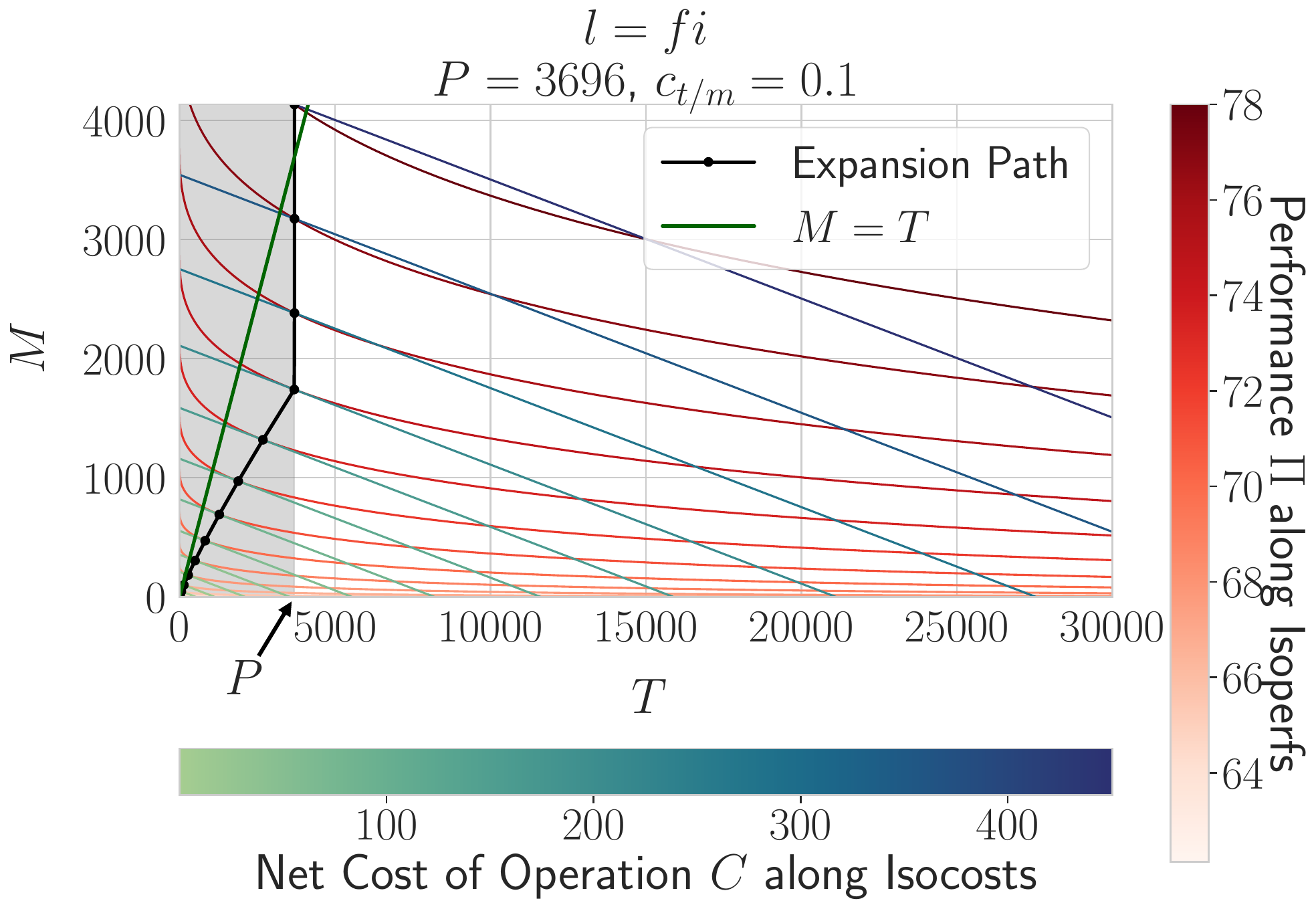}
    \caption{}
    \label{}
    \end{subfigure}%
    \begin{subfigure}[t]{0.45\textwidth}
    \includegraphics[width=0.95\textwidth]{{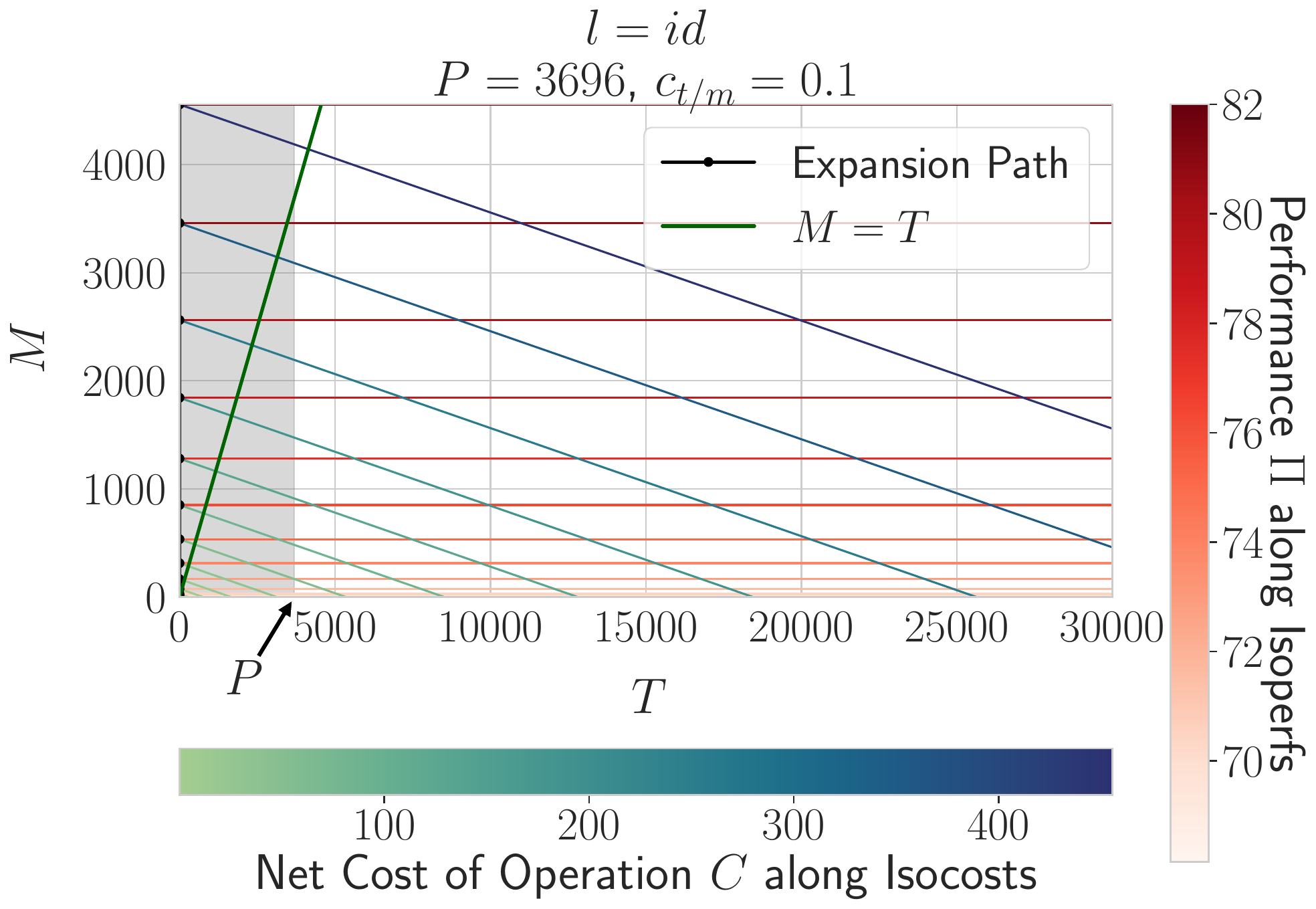}}
    \caption{}
    \label{}
    \end{subfigure}
    \begin{subfigure}[t]{0.45\textwidth}
    \includegraphics[width=0.95\textwidth]{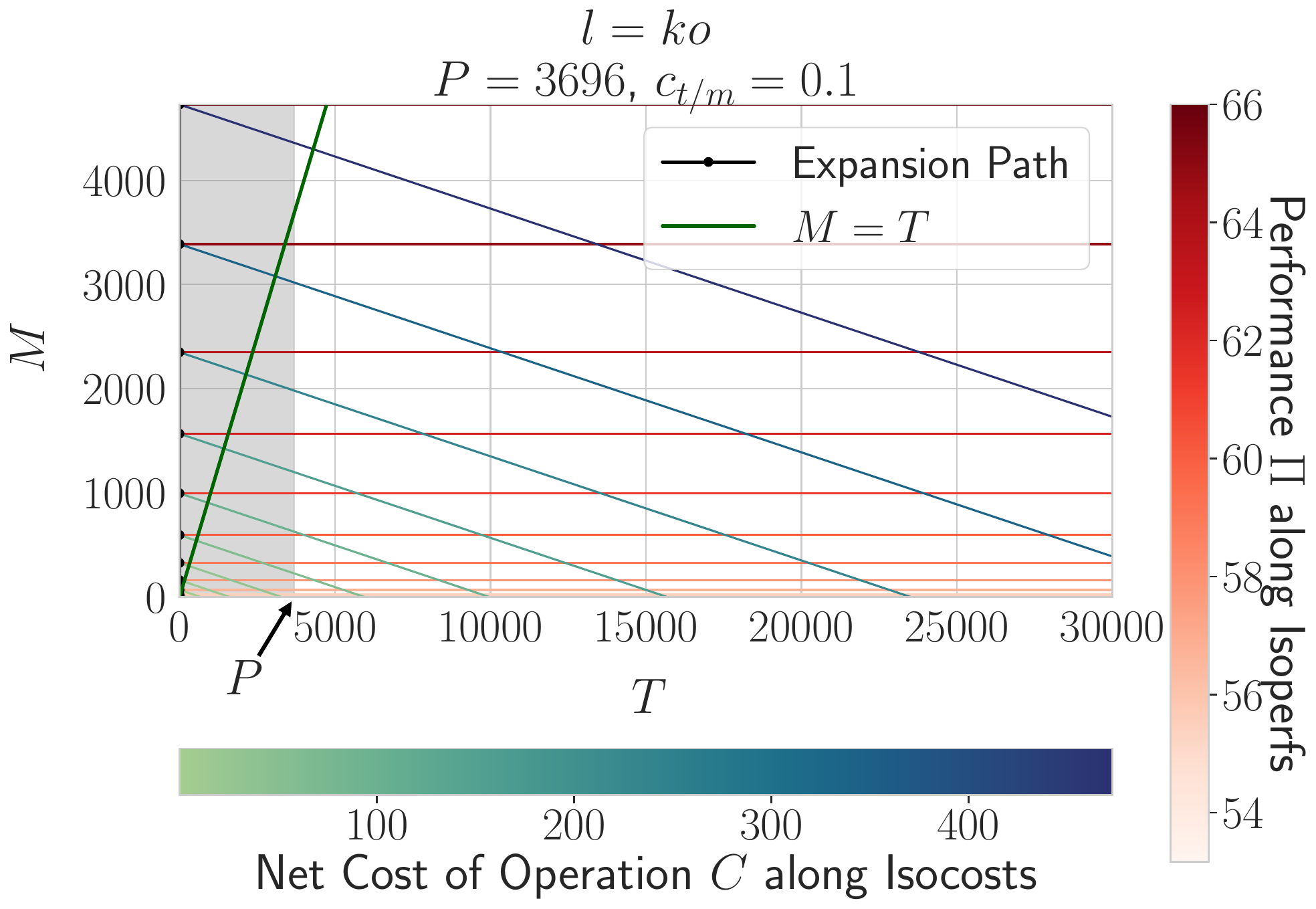}
    \caption{}
    \label{}
    \end{subfigure}%
    \begin{subfigure}[t]{0.45\textwidth}
    \includegraphics[width=0.95\textwidth]{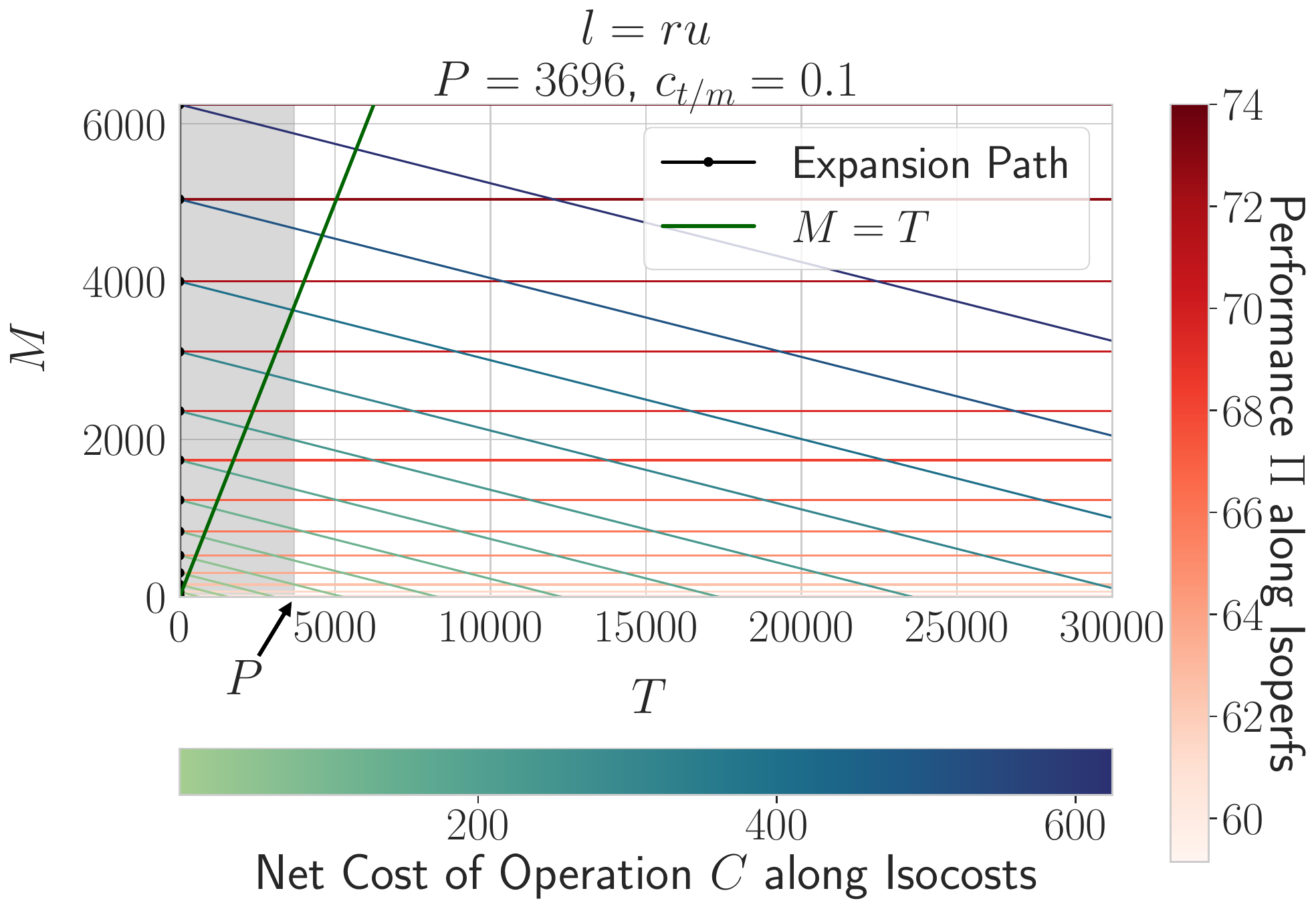}
    \caption{}
    \label{}
    \end{subfigure}
    \begin{subfigure}[t]{0.45\textwidth}
    \includegraphics[width=0.95\textwidth]{{figures/exp_paths/expansion_path_sw_3696_01_wtbg.pdf}}
    \caption{}
    \label{}
    \end{subfigure}%
    \begin{subfigure}[t]{0.45\textwidth}
    \includegraphics[width=0.95\textwidth]{{figures/exp_paths/expansion_path_te_3696_01_wtbg.pdf}}
    \caption{}
    \label{}
    \end{subfigure}
    \caption{M-T diagrams for different languages for $P = 3696$ and $c_{t/m}$ = 0.1}
    \label{fig:p3kc01}
\end{figure*}

\begin{figure*}
    \centering
    \begin{subfigure}[t]{0.45\textwidth}
    \includegraphics[width=0.95\textwidth]{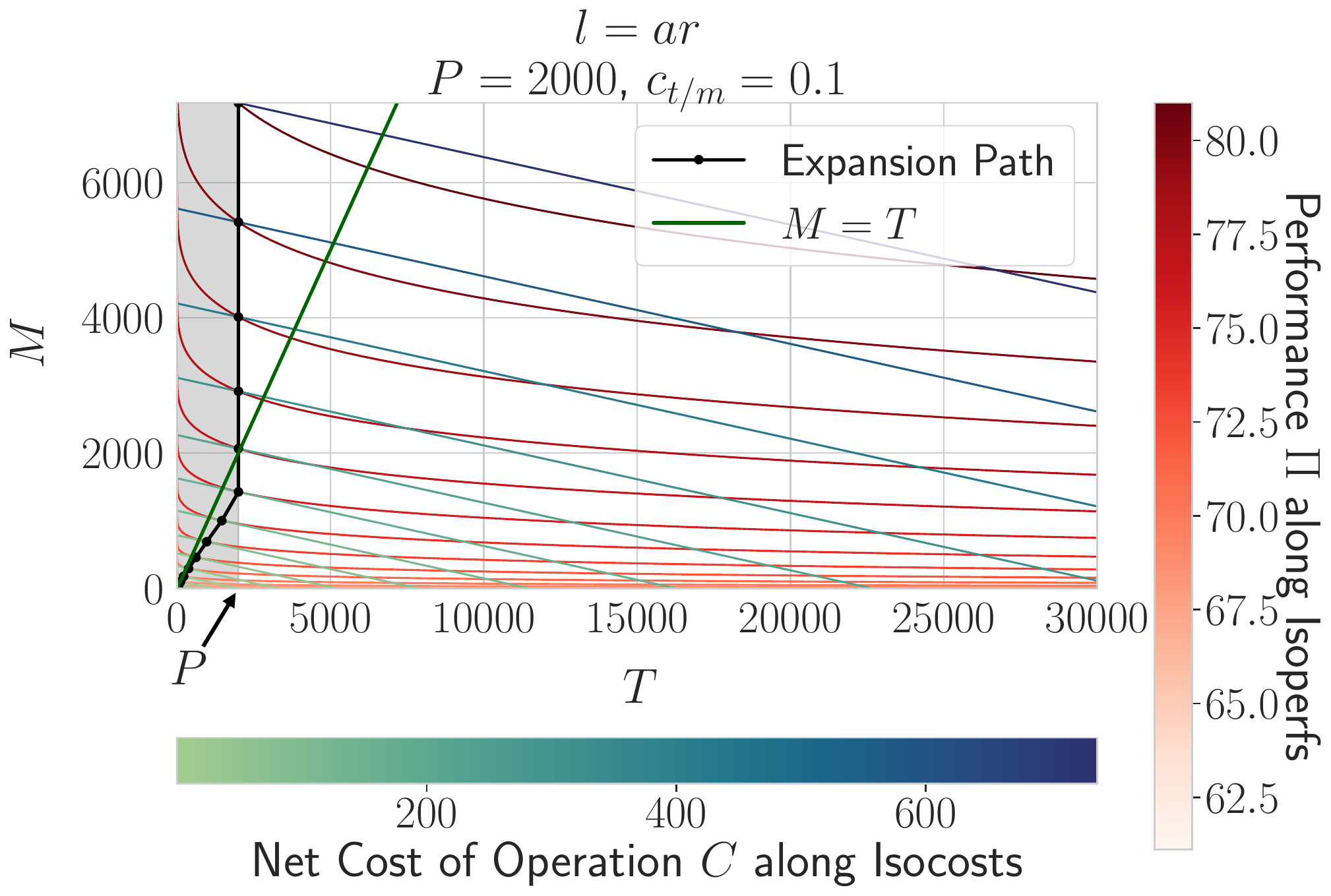}
    \caption{}
    \label{}
    \end{subfigure}%
    \begin{subfigure}[t]{0.45\textwidth}
    \includegraphics[width=0.95\textwidth]{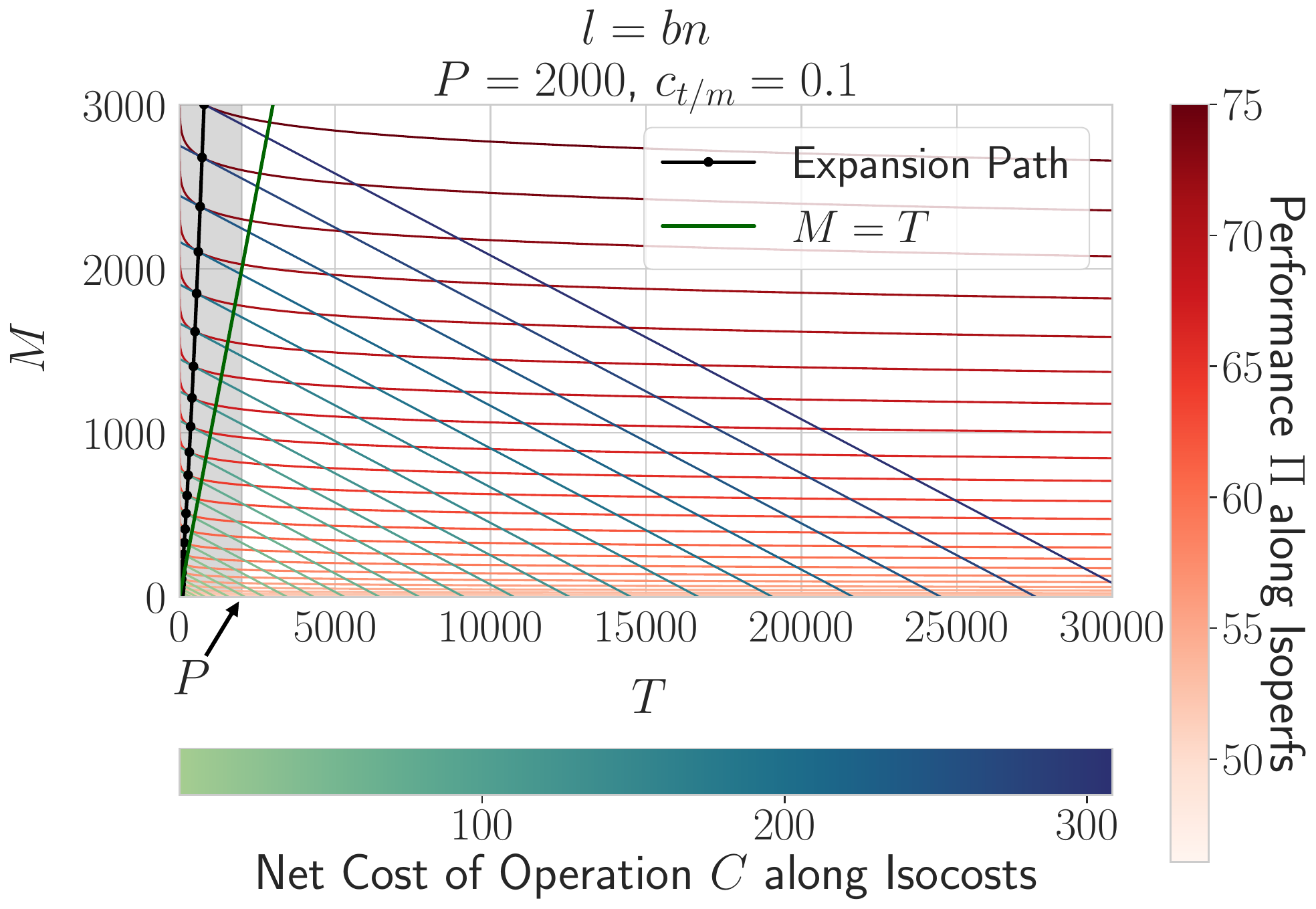}
    \caption{}
    \label{}
    \end{subfigure}
    \begin{subfigure}[t]{0.45\textwidth}
    \includegraphics[width=0.95\textwidth]{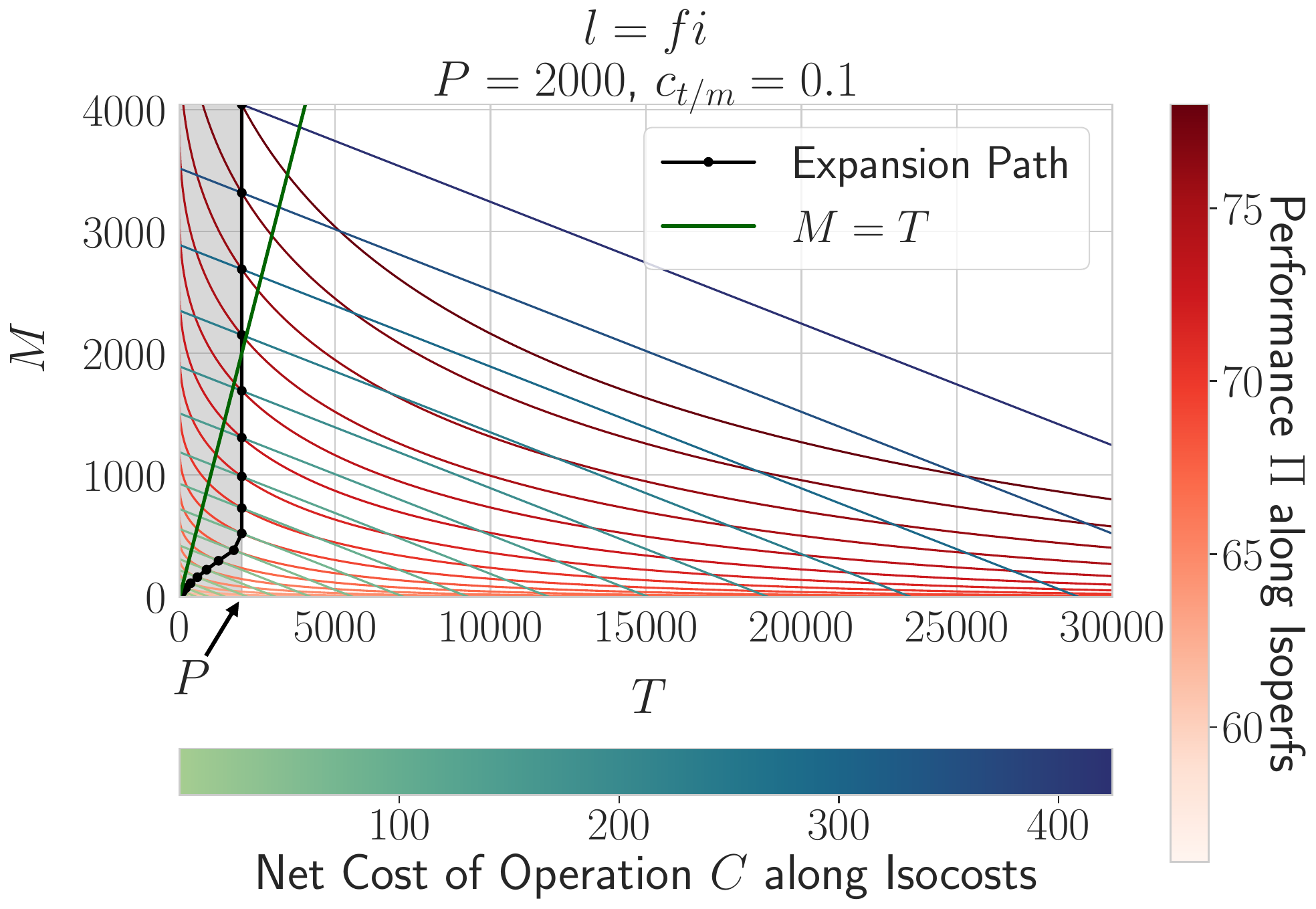}
    \caption{}
    \label{}
    \end{subfigure}%
    \begin{subfigure}[t]{0.45\textwidth}
    \includegraphics[width=0.95\textwidth]{{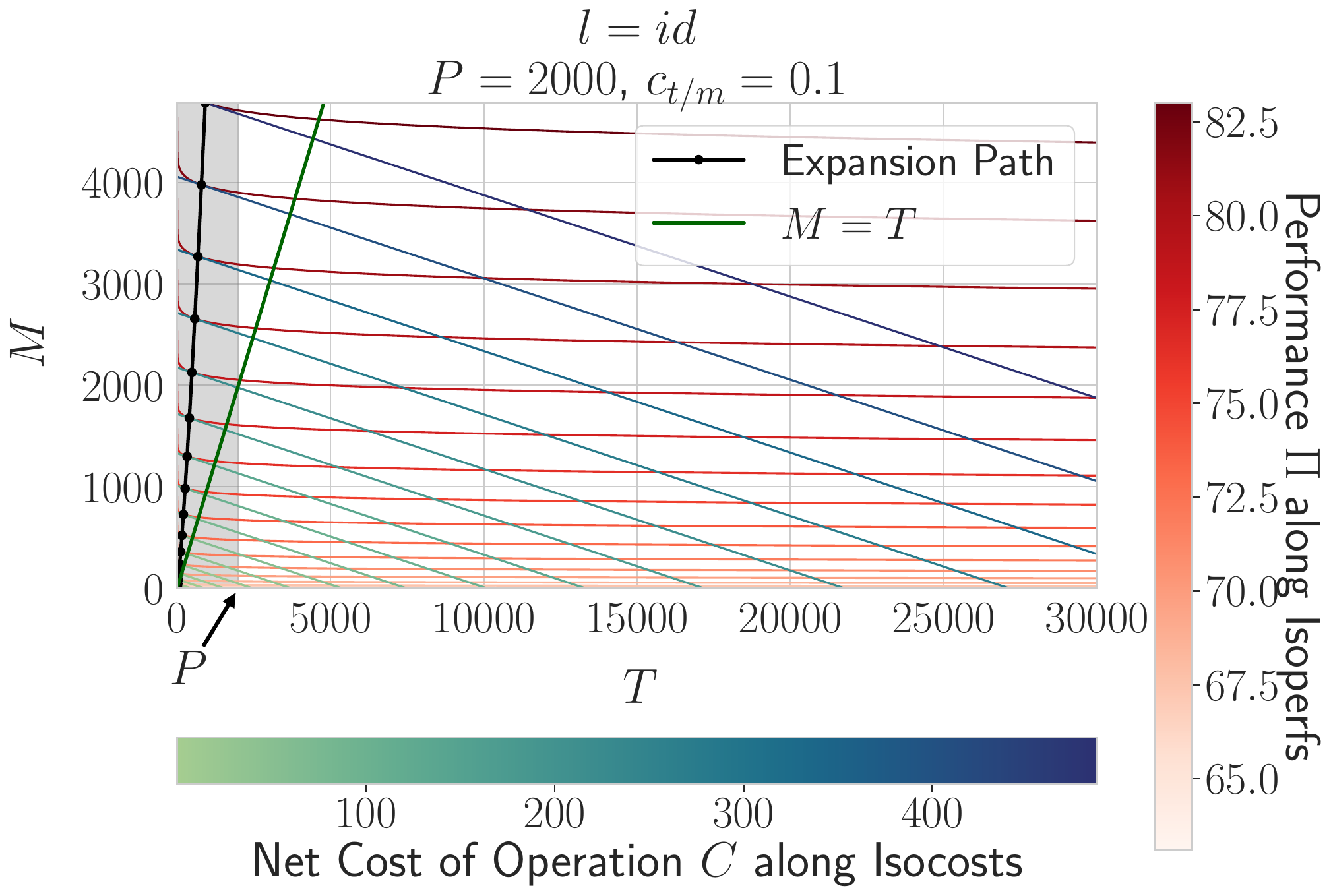}}
    \caption{}
    \label{}
    \end{subfigure}
    \begin{subfigure}[t]{0.45\textwidth}
    \includegraphics[width=0.95\textwidth]{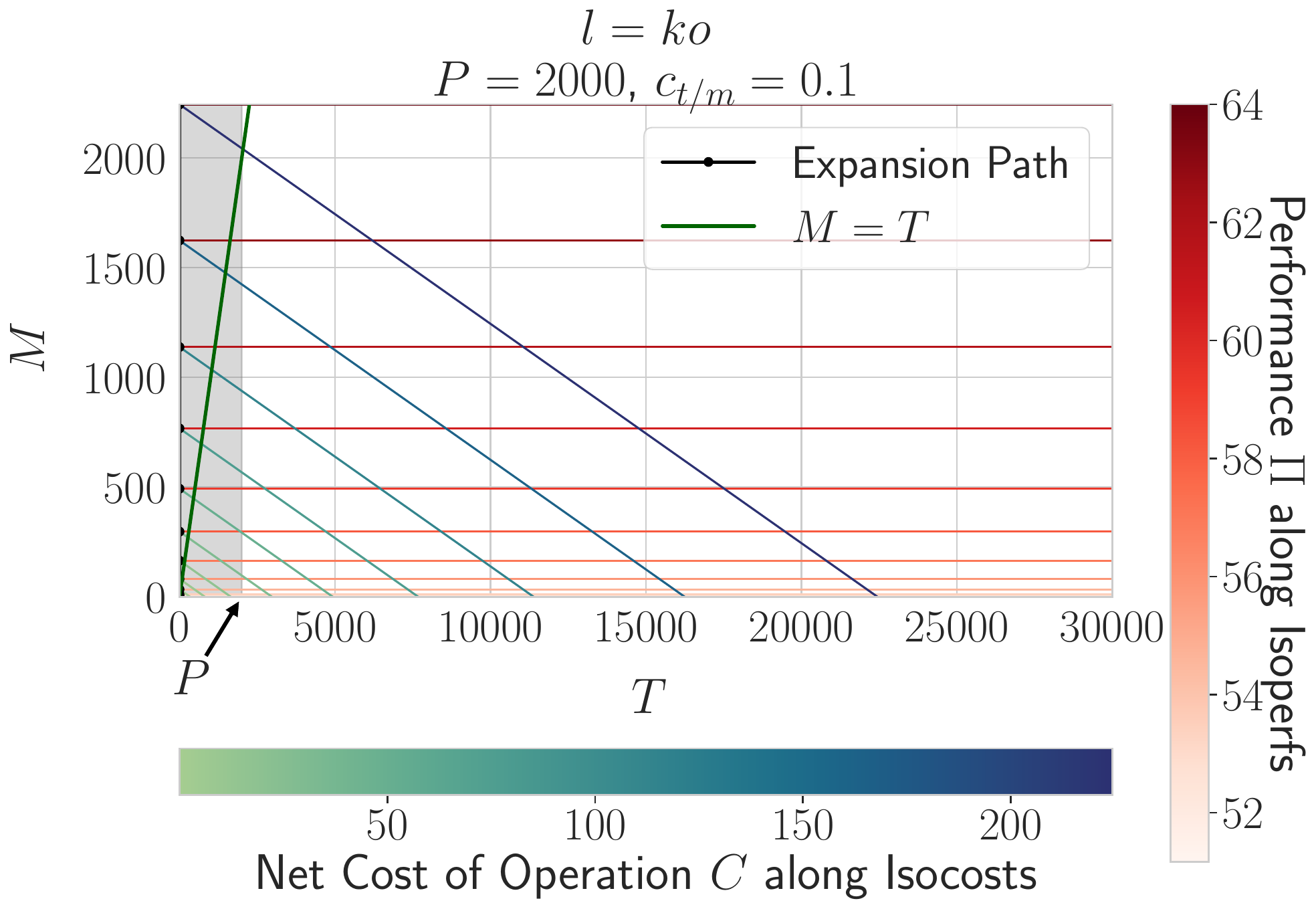}
    \caption{}
    \label{}
    \end{subfigure}%
    \begin{subfigure}[t]{0.45\textwidth}
    \includegraphics[width=0.95\textwidth]{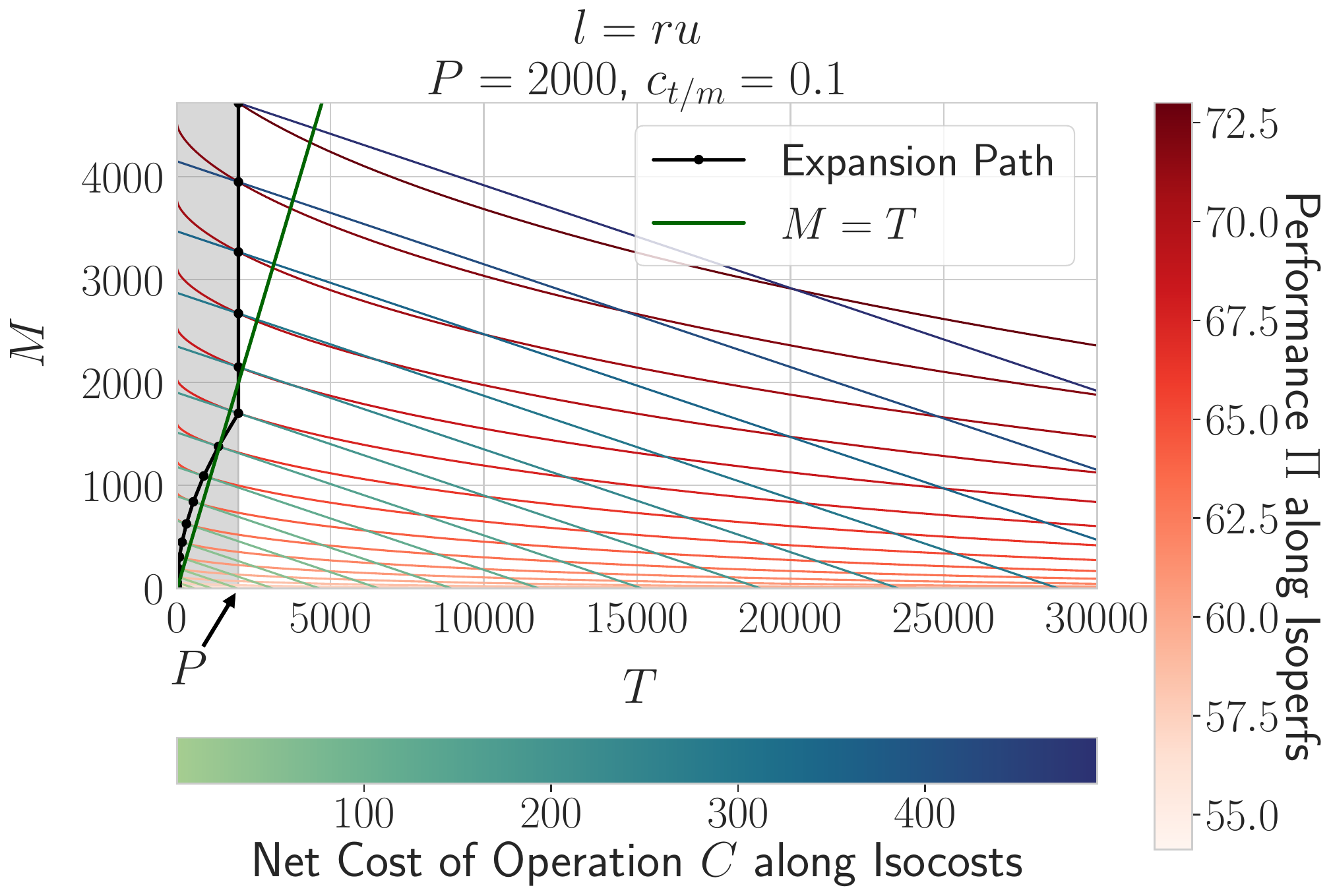}
    \caption{}
    \label{}
    \end{subfigure}
    \begin{subfigure}[t]{0.45\textwidth}
    \includegraphics[width=0.95\textwidth]{{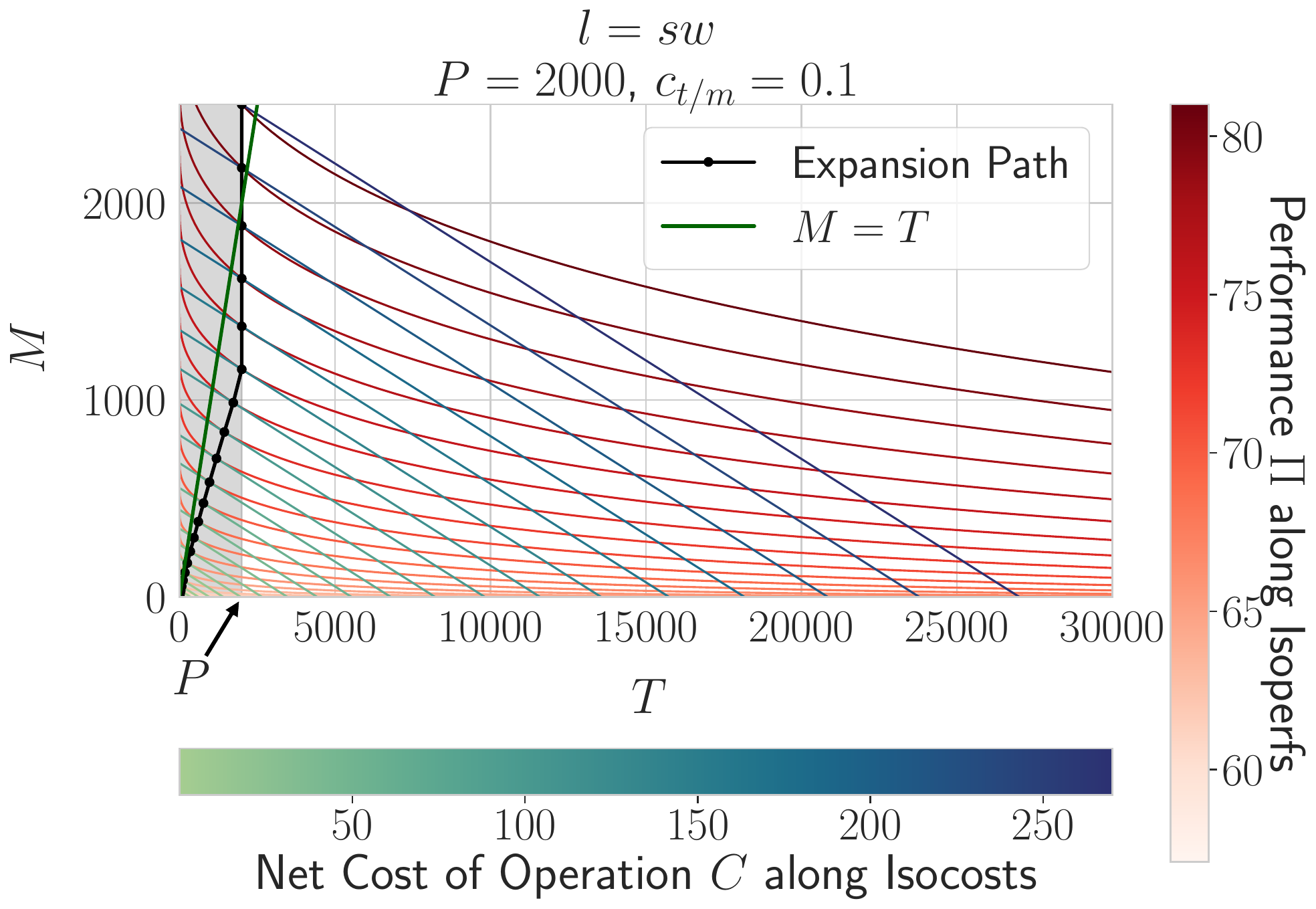}}
    \caption{}
    \label{}
    \end{subfigure}%
    \begin{subfigure}[t]{0.45\textwidth}
    \includegraphics[width=0.95\textwidth]{{figures/exp_paths/expansion_path_te_2000_01_wtbg.pdf}}
    \caption{}
    \label{}
    \end{subfigure}
    \caption{M-T diagrams for different languages for $P = 2000$ and $c_{t/m}$ = 0.1}
    \label{fig:p2kc01}
\end{figure*}

\begin{figure*}
    \centering
    \begin{subfigure}[t]{0.45\textwidth}
    \includegraphics[width=0.95\textwidth]{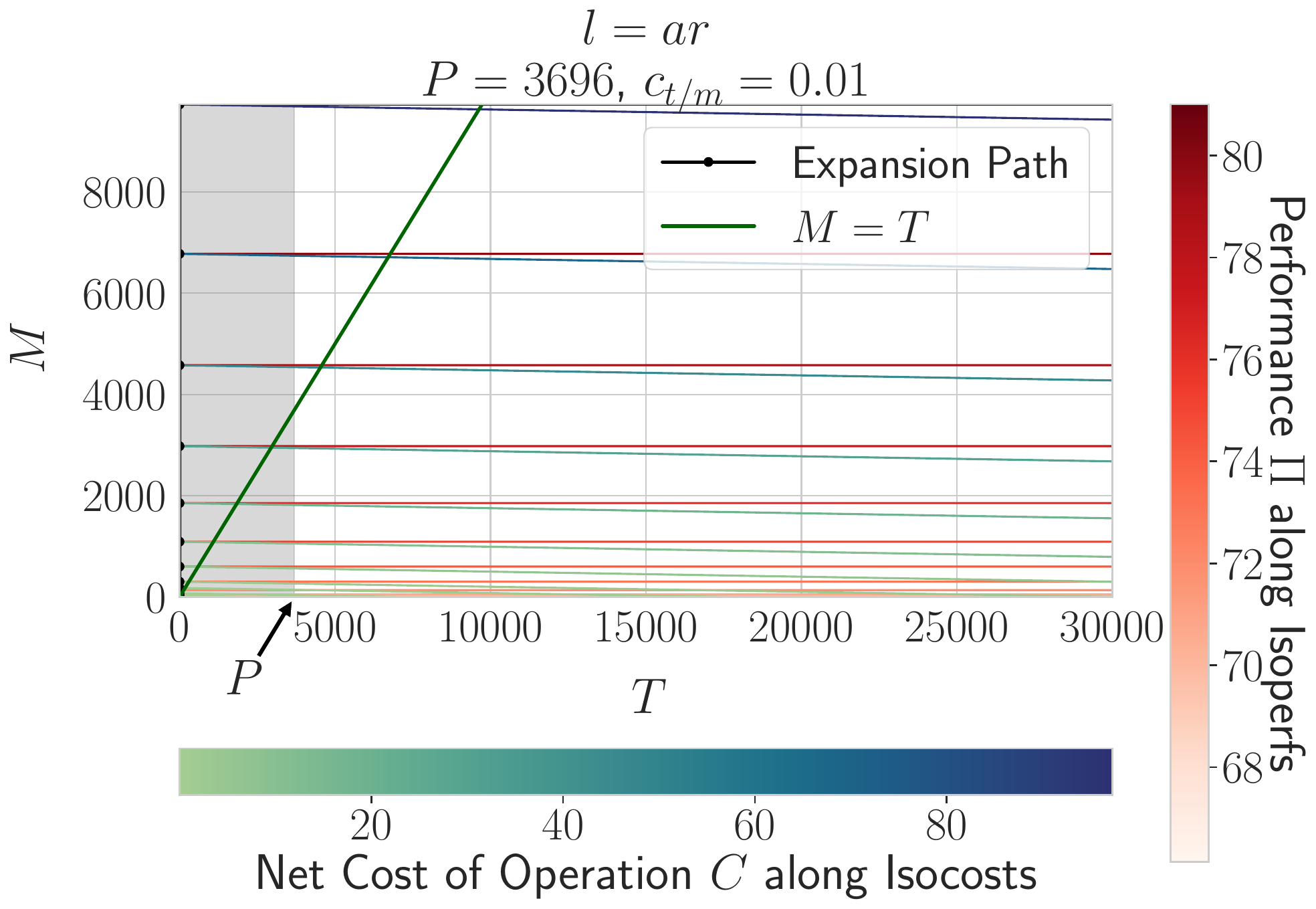}
    \caption{}
    \label{}
    \end{subfigure}%
    \begin{subfigure}[t]{0.45\textwidth}
    \includegraphics[width=0.95\textwidth]{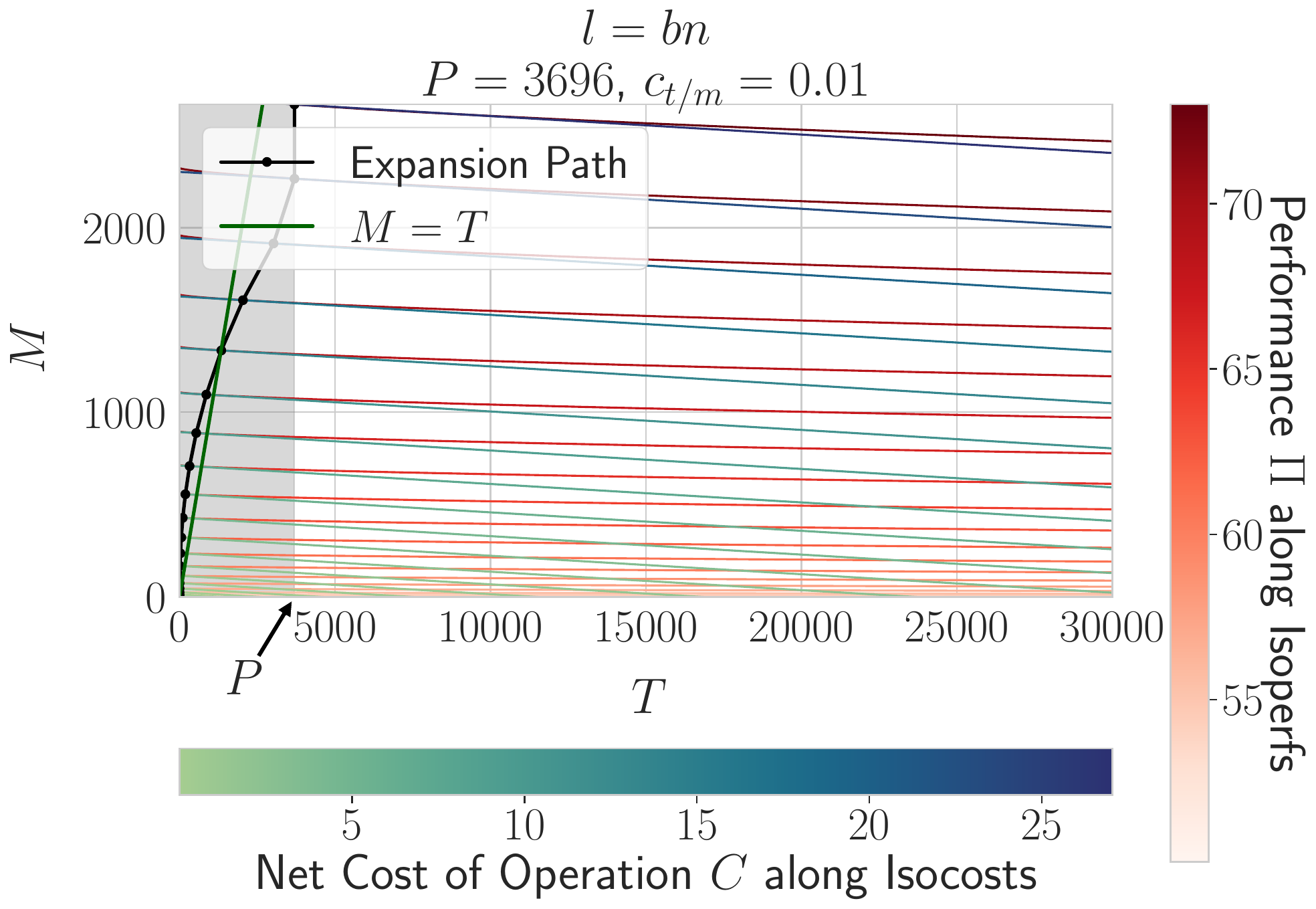}
    \caption{}
    \label{}
    \end{subfigure}
    \begin{subfigure}[t]{0.45\textwidth}
    \includegraphics[width=0.95\textwidth]{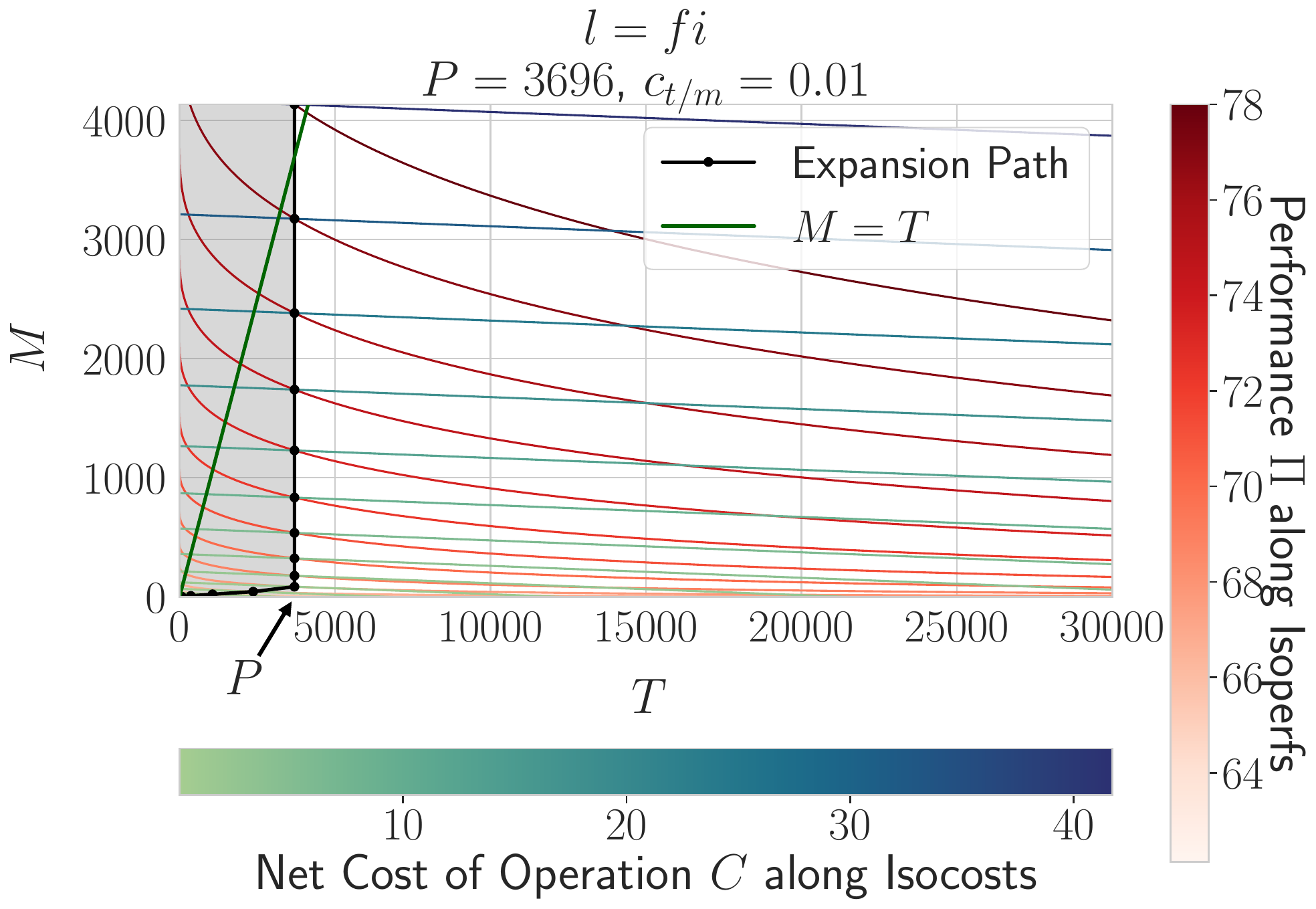}
    \caption{}
    \label{}
    \end{subfigure}%
    \begin{subfigure}[t]{0.45\textwidth}
    \includegraphics[width=0.95\textwidth]{{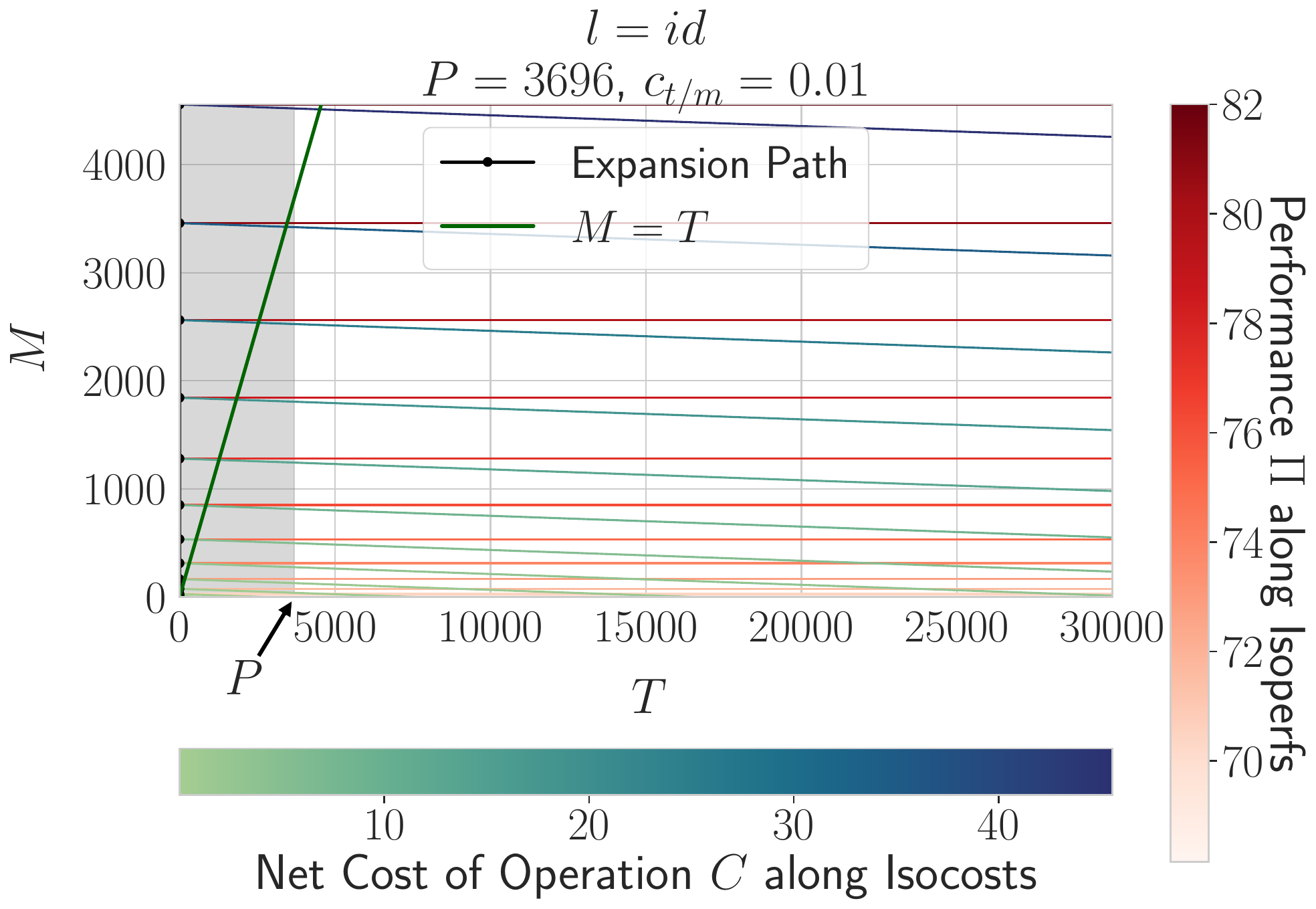}}
    \caption{}
    \label{}
    \end{subfigure}
    \begin{subfigure}[t]{0.45\textwidth}
    \includegraphics[width=0.95\textwidth]{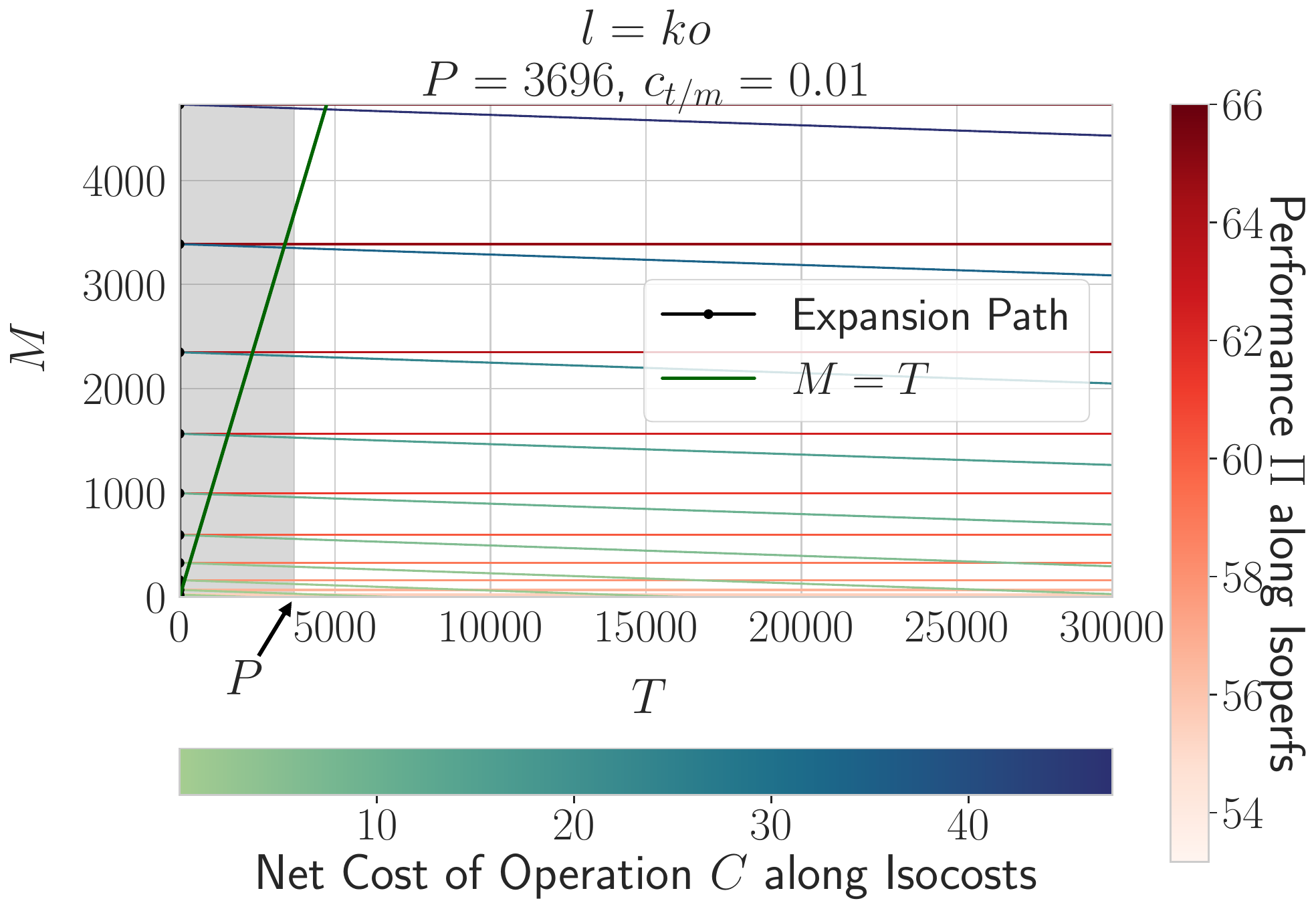}
    \caption{}
    \label{}
    \end{subfigure}%
    \begin{subfigure}[t]{0.45\textwidth}
    \includegraphics[width=0.95\textwidth]{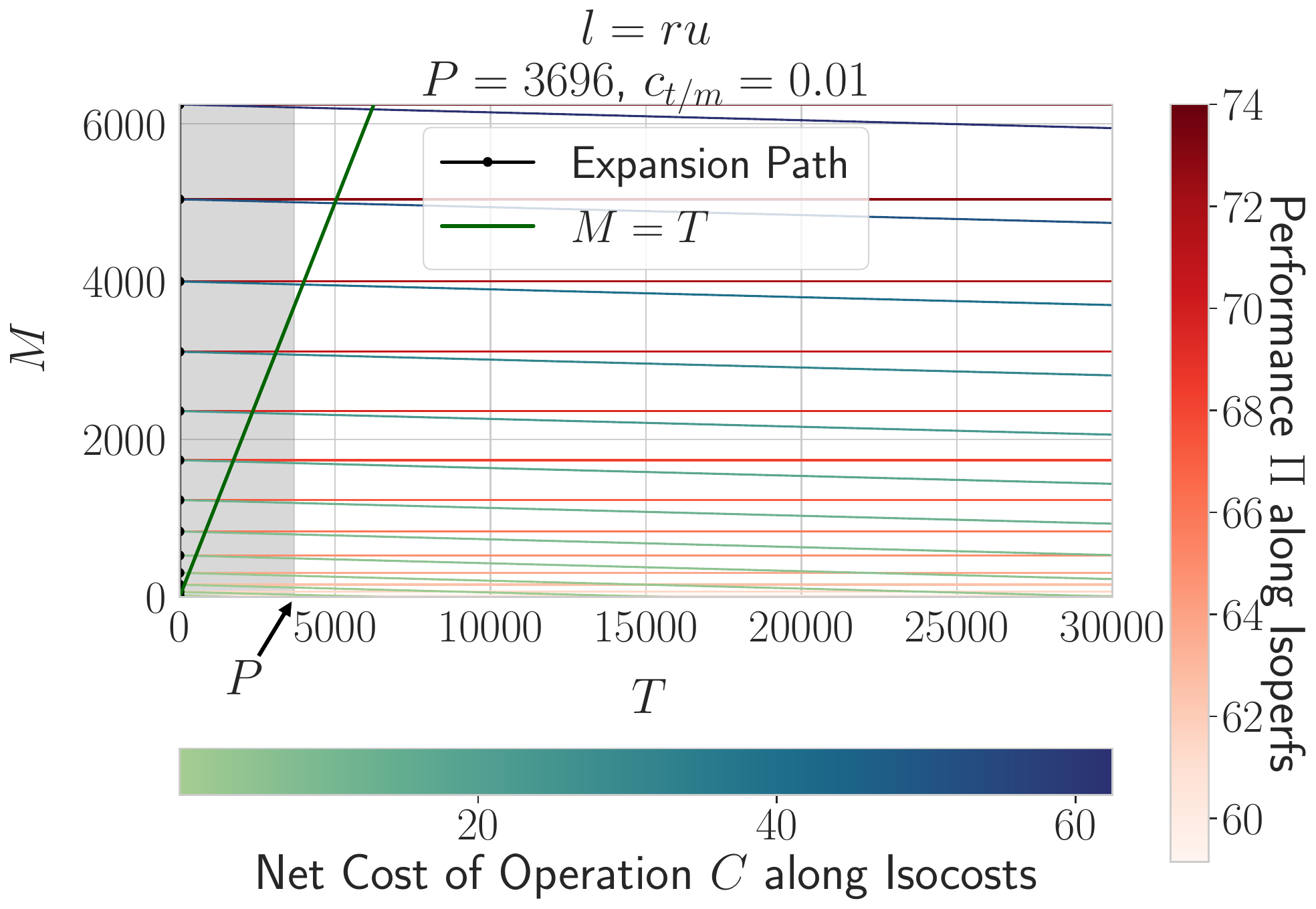}
    \caption{}
    \label{}
    \end{subfigure}
    \begin{subfigure}[t]{0.45\textwidth}
    \includegraphics[width=0.95\textwidth]{{figures/exp_paths/expansion_path_sw_3696_001_wtbg.pdf}}
    \caption{}
    \label{}
    \end{subfigure}%
    \begin{subfigure}[t]{0.45\textwidth}
    \includegraphics[width=0.95\textwidth]{{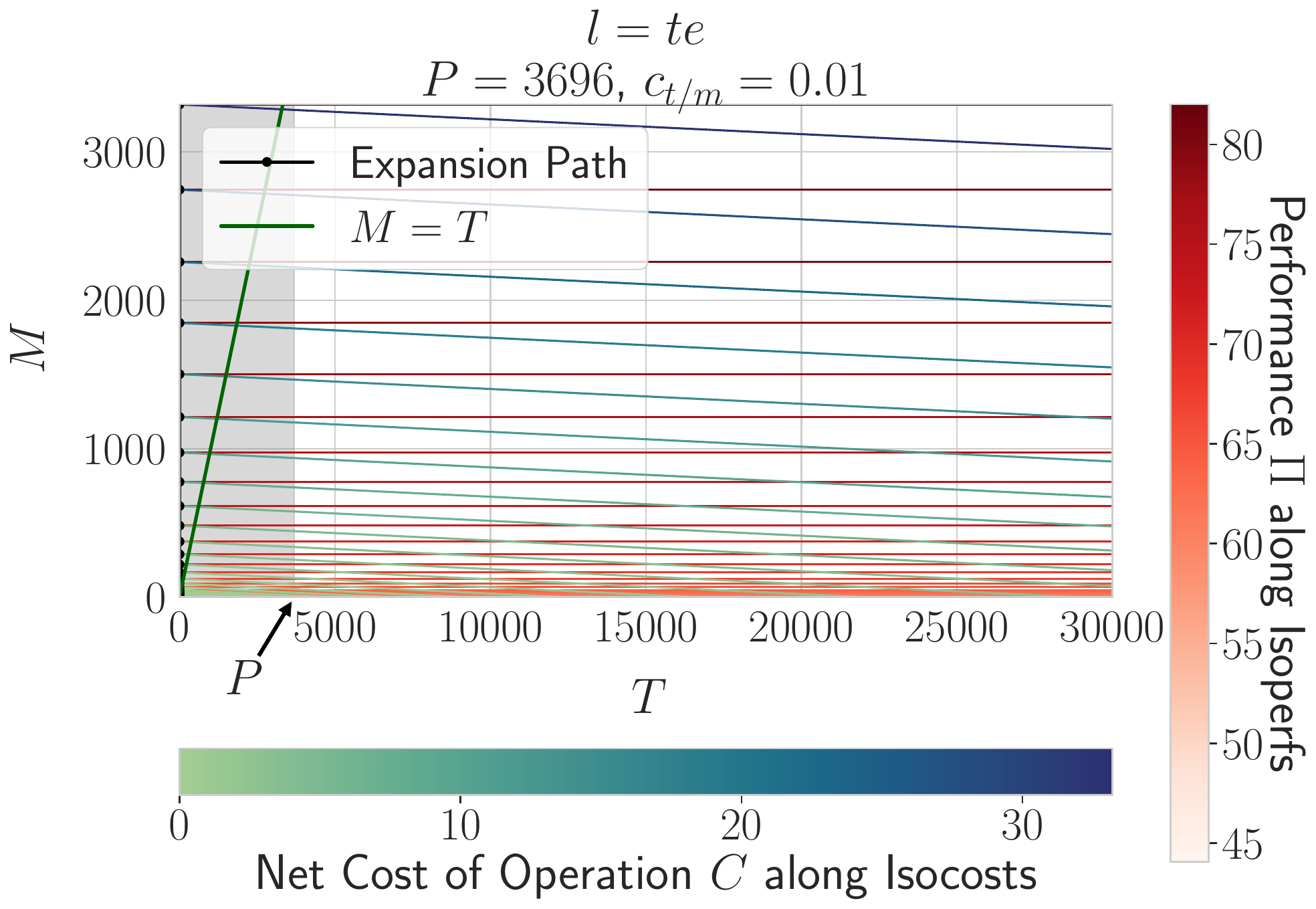}}
    \caption{}
    \label{}
    \end{subfigure}
    \caption{M-T diagrams for different languages for $P = 3696$ and $c_{t/m}$ = 0.01}
    \label{fig:p3kc001}
\end{figure*}

\begin{figure*}
    \centering
    \begin{subfigure}[t]{0.45\textwidth}
    \includegraphics[width=0.95\textwidth]{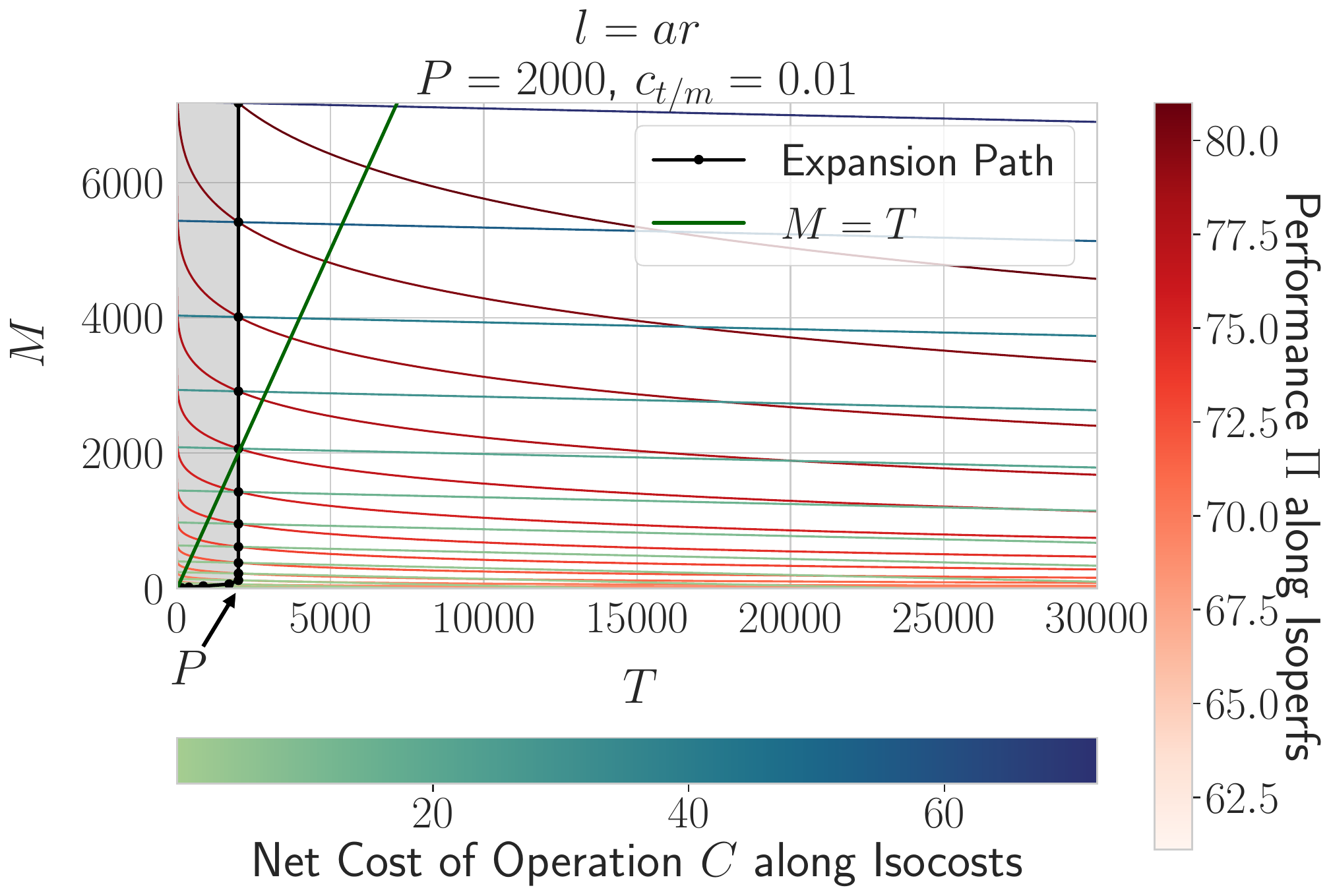}
    \caption{}
    \label{}
    \end{subfigure}%
    \begin{subfigure}[t]{0.45\textwidth}
    \includegraphics[width=0.95\textwidth]{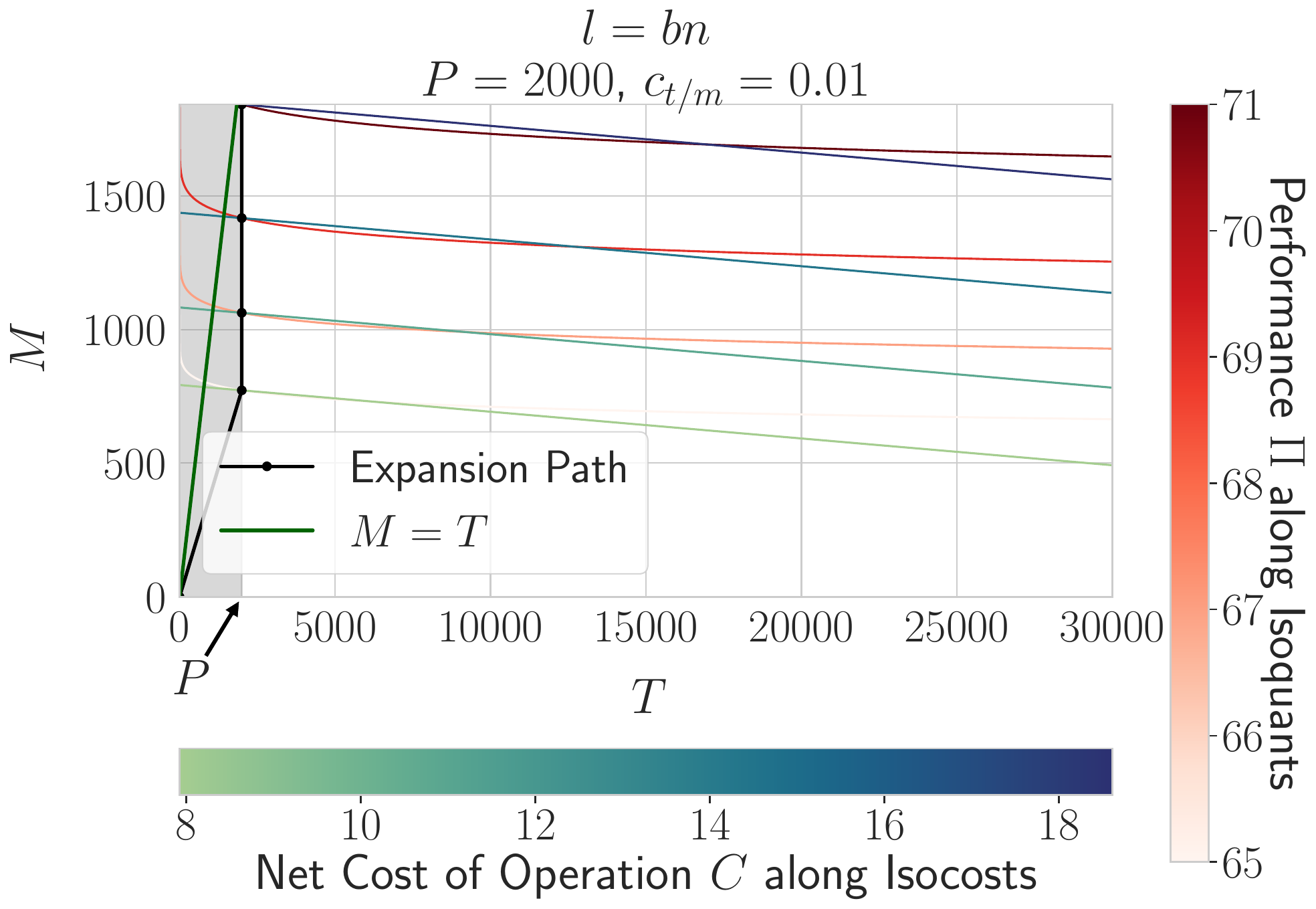}
    \caption{}
    \label{}
    \end{subfigure}
    \begin{subfigure}[t]{0.45\textwidth}
    \includegraphics[width=0.95\textwidth]{{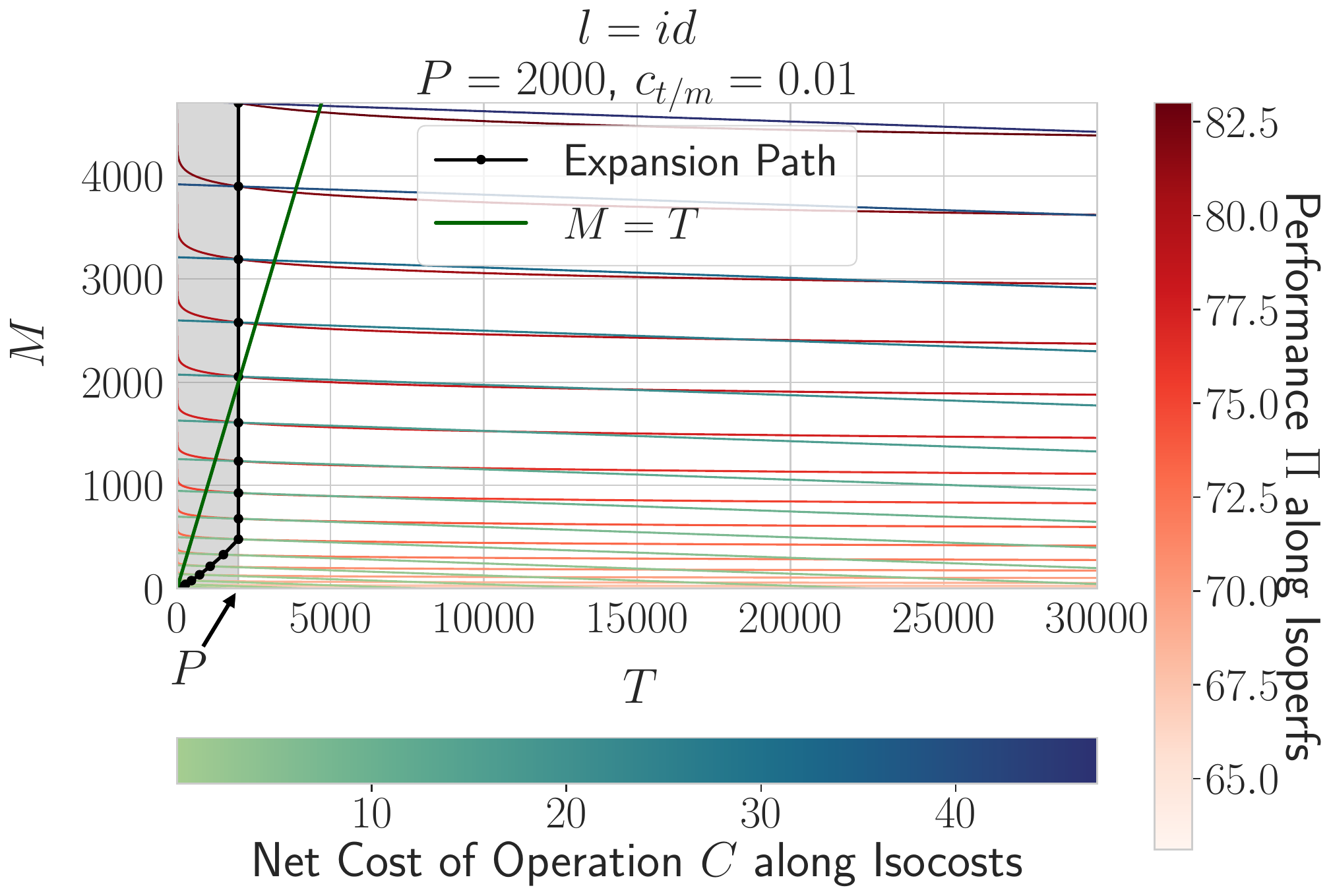}}
    \caption{}
    \label{}
    \end{subfigure}%
    \begin{subfigure}[t]{0.45\textwidth}
    \includegraphics[width=0.95\textwidth]{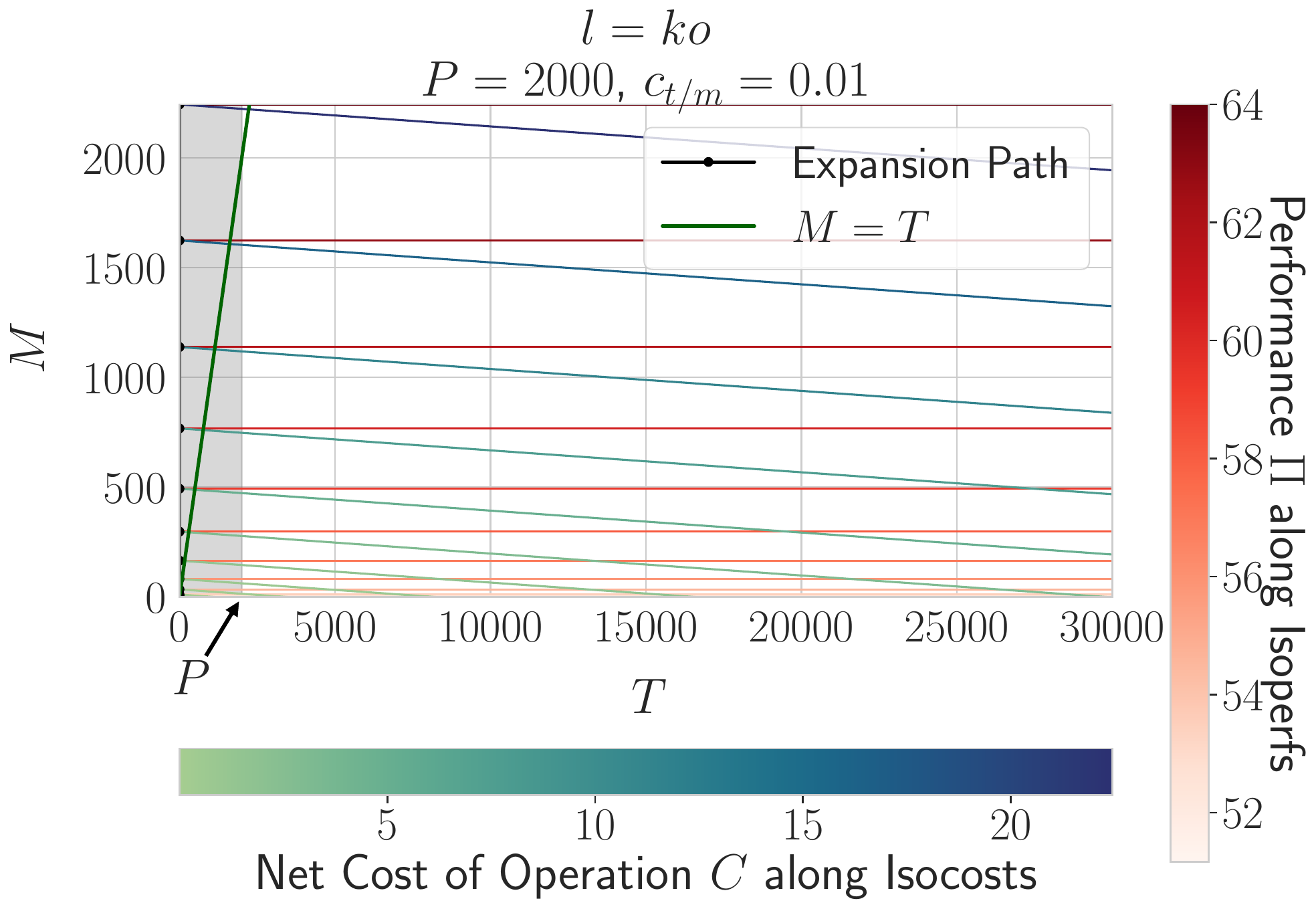}
    \caption{}
    \label{}
    \end{subfigure}
    \begin{subfigure}[t]{0.45\textwidth}
    \includegraphics[width=0.95\textwidth]{{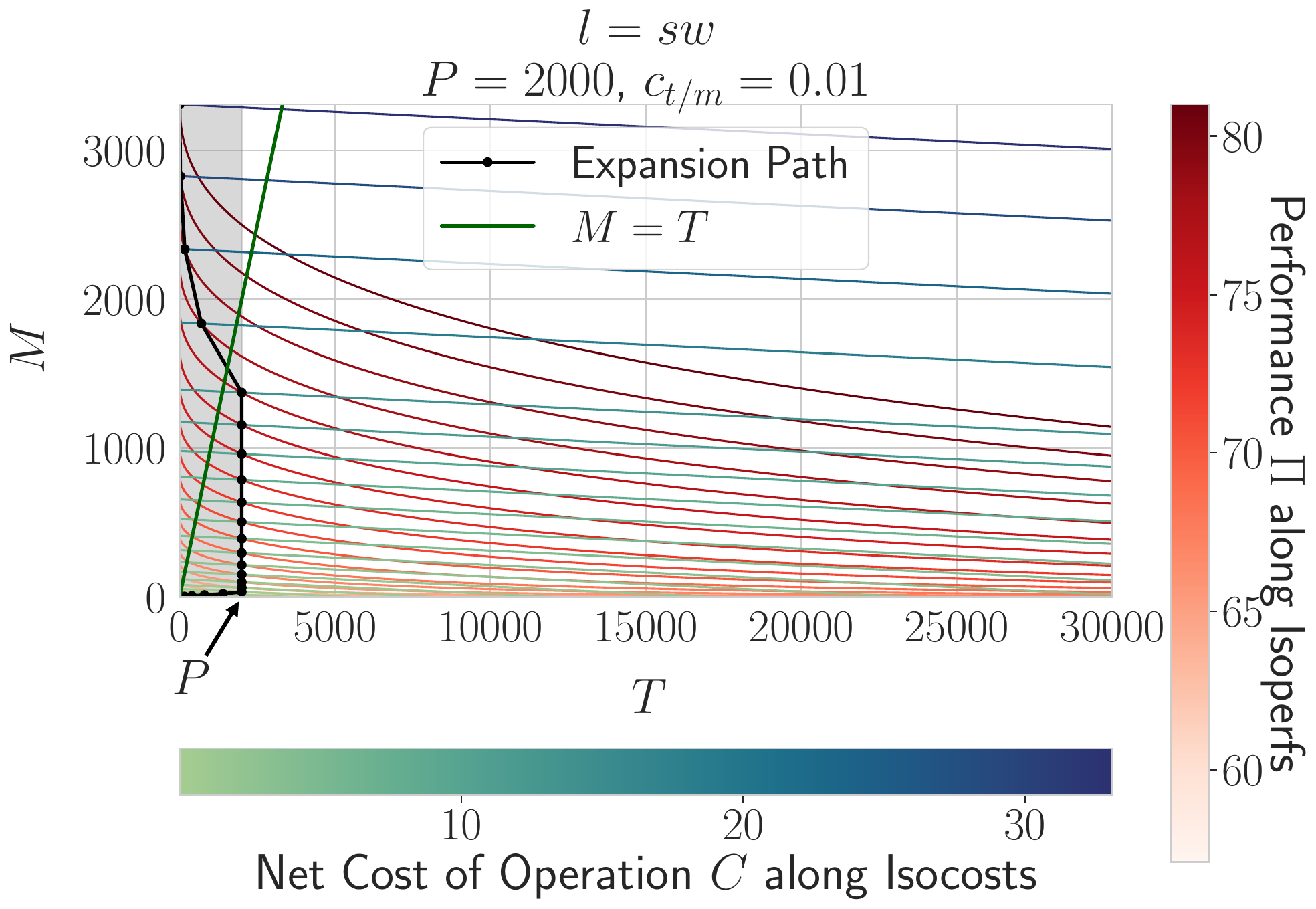}}
    \caption{}
    \label{}
    \end{subfigure}%
    \begin{subfigure}[t]{0.45\textwidth}
    \includegraphics[width=0.95\textwidth]{{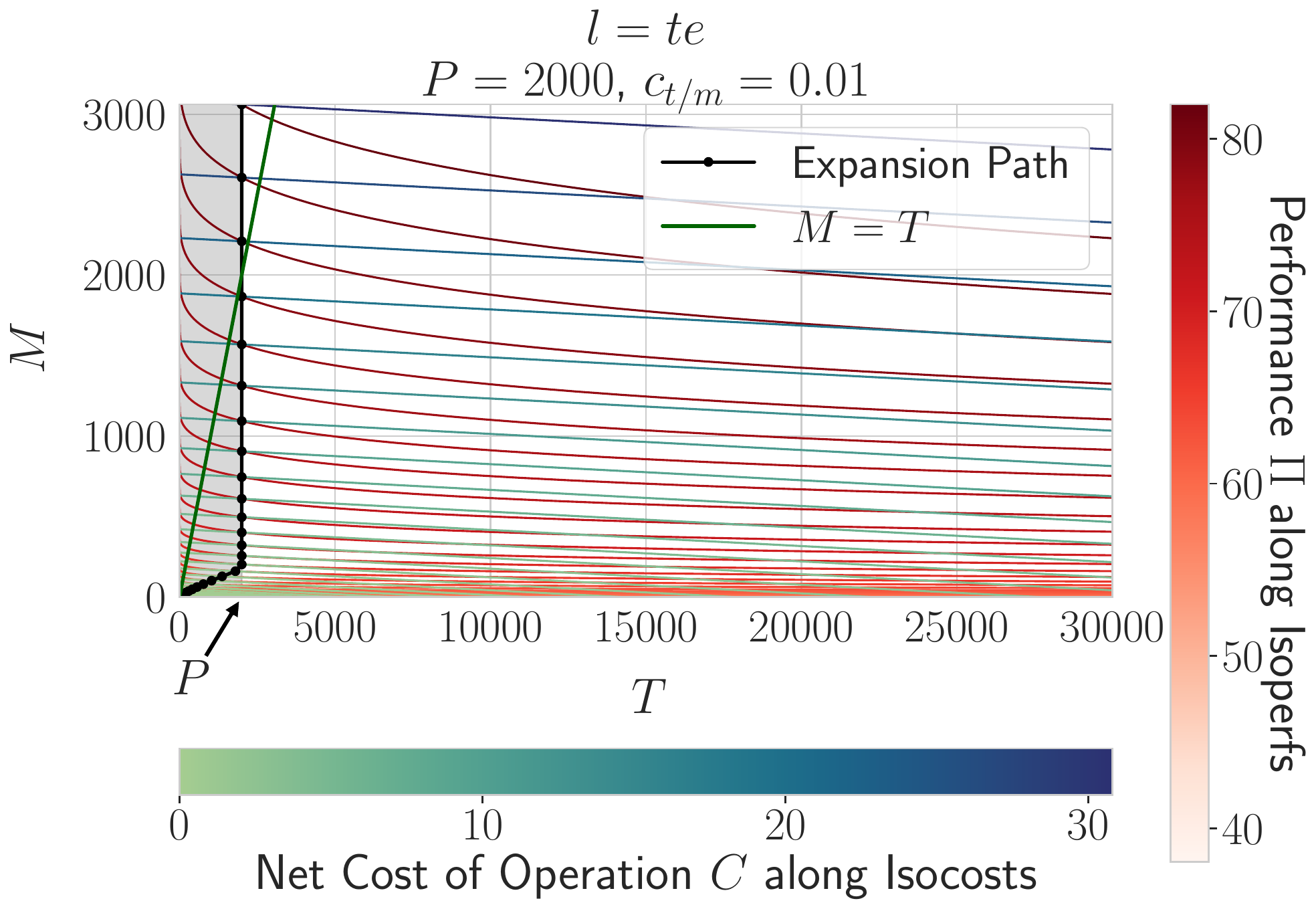}}
    \caption{}
    \label{}
    \end{subfigure}
    \caption{M-T diagrams for different languages for $P = 2000$ and $c_{t/m}$ = 0.01}
    \label{fig:p2kc001}
\end{figure*}

\begin{figure}
    \centering
    \includegraphics[width=0.45\textwidth]{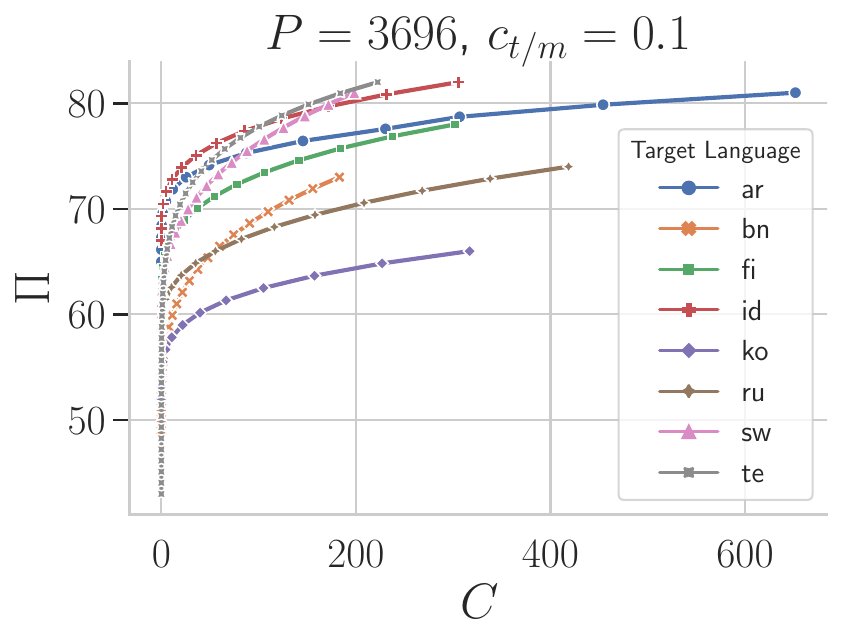}
    \caption{Performance vs the minimum costs for different languages for $c = 0.1$. As expected the overall costs are now lower than in figure \ref{fig:cvspi}, since the manual data is cheaper in this case.}
    \label{fig:cvspi_c01}
\end{figure}

\end{document}